\documentclass[lettersize,journal]{IEEEtran}
\usepackage{amsmath,amsfonts}
\usepackage{algorithmic}
\usepackage{algorithm}
\usepackage{array}
\usepackage[caption=false,font=normalsize,labelfont=sf,textfont=sf]{subfig}
\usepackage{textcomp}
\usepackage{stfloats}
\usepackage{url}
\usepackage{verbatim}
\usepackage{graphicx}
\usepackage{cite}
\usepackage{CJKutf8}
\usepackage{subfig}
\usepackage{float}
\makeatletter
 \let\NAT@parse\undefined
 \makeatother
\usepackage[pdftex,
        colorlinks=true,
        bookmarks=true,
        citecolor=red,
        linkcolor=blue,
        pagebackref=true,
        pdfstartview=Fit,
        breaklinks=true
]{hyperref}
\usepackage{cleveref}
\begin{document}
\newcommand{\blue}[1]{{\color{blue}#1}}
\newcommand{\red}[1]{{\color{red}{\sout{#1}}}}
\newcommand{\etal}{\textit{et al.~}}
\def\etc{\textit{etc.~}}  
\def\eg{\textit{e.g.},~}
\def\ie{\textit{i.e.},~}

\title{A Haptic Robot Finger Designed for Guqin Instrument Playing }
\author{Tianwei Zhang$^{1,2}$, Hanming Yan$^{3}$, Yang Yang$^{1,4}$, Ziya Wang$^{3*}$
\thanks{$^{1}$The Shenzhen Institute of Artificial Intelligence and Robotics for Society, Shenzhen, China}
\thanks{{$^2$}The Chinese University of Hong Kong - Shenzhen, Shenzhen, China}
\thanks{{$^3$}School of Physics and Optoelectronic Engineering, Shenzhen University, Shenzhen, China}
\thanks{{$^4$}Shenzhen International Graduate School, Tsinghua University, Shenzhen, China}

\thanks{This work was supported by the Shenzhen Science and Technology Program Grant No. JSGGKQTD20221101115656029, ZDCY20250901094531003 and KJZD20230923113801004; this work was also supported by Linkerbot Beijing Technology Co., Ltd.}
\thanks{* Corresponding Author: \tt\small {wangziya@szu.edu.cn}}
}
\maketitle

\begin{abstract}
With the rapid advancement of humanoid robotics and embodied intelligence technologies, numerous musical instrument-playing robots have emerged in recent years, such as pianos, chime bells, and taiko drums. These robots primarily employ open-loop positional control, rendering them incapable of operating instruments requiring dexterous hands and precise tactile perception, such as a violin, guitar, and guqin. 
This paper describes the design and validation of a high-precision tactile-sensing finger. By mimicking the shape of the fingertip and fingernail found on a human finger, we develop a biomimetic multimodal haptic fingertip and validate it on selected guqin string-contact tasks, including open-string and stopped-note comparisons, harmonic-tuning, and tactile-triggered bimanual coordination, using the guqin, a traditional Chinese musical instrument, as a challenging validation scenario rather than as a fully demonstrated robotic performance system.
This research integrates tactile sensing with robotics technology, thereby contributing to applications in world heritage conservation and cultural dissemination.
\end{abstract}

\begin{CJK}{UTF8}{gbsn}

\section{Introduction}

\begin{figure}[t]
\captionsetup[subfloat]{labelformat=empty}
  \subfloat[\scriptsize{(a)  The designed haptic finger for robot guqin playing}]{\includegraphics[width=\columnwidth]{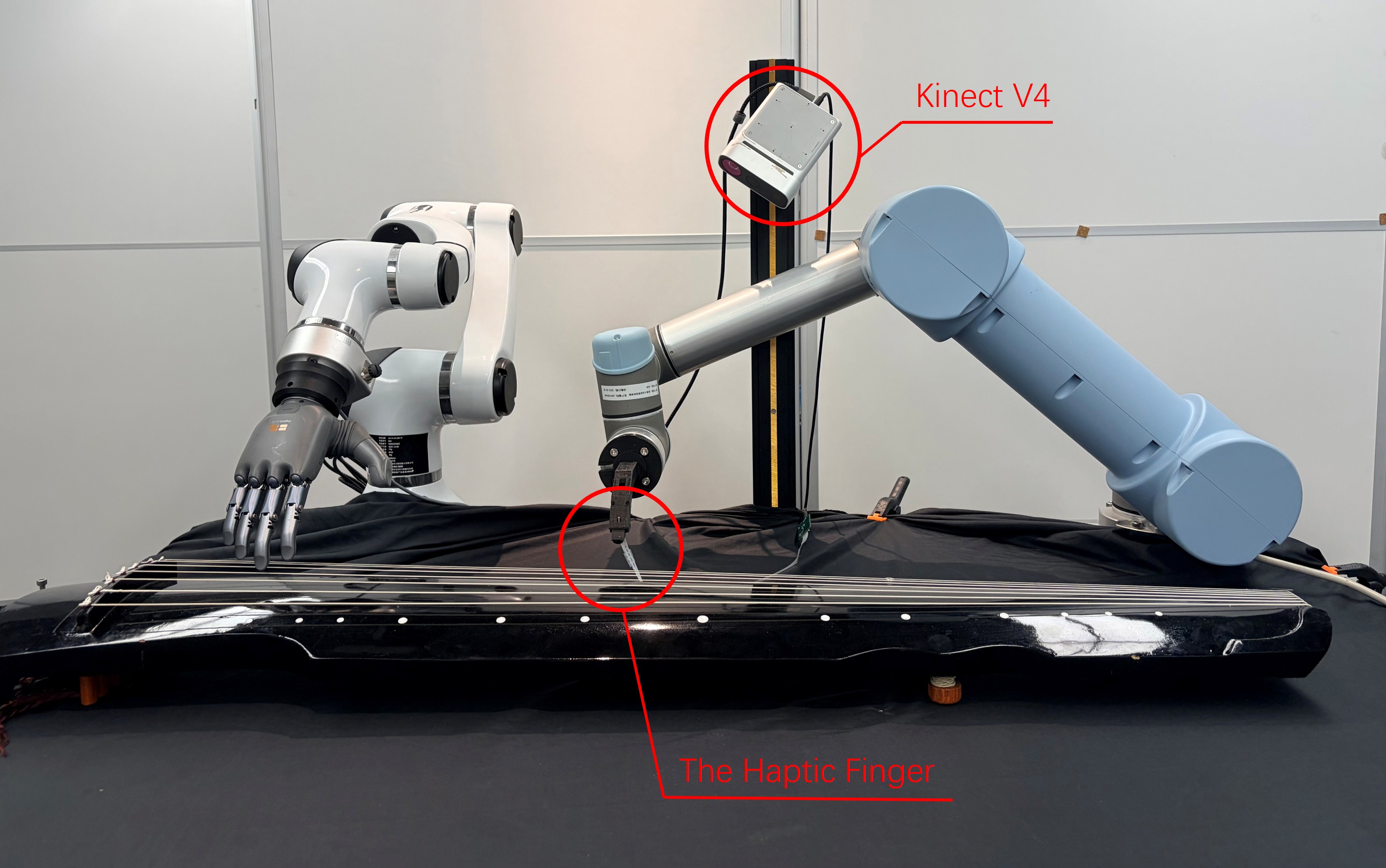}}
\\
\subfloat[\scriptsize{(b) The structure of the guqin and its musical notation}]{\includegraphics[width=\columnwidth]{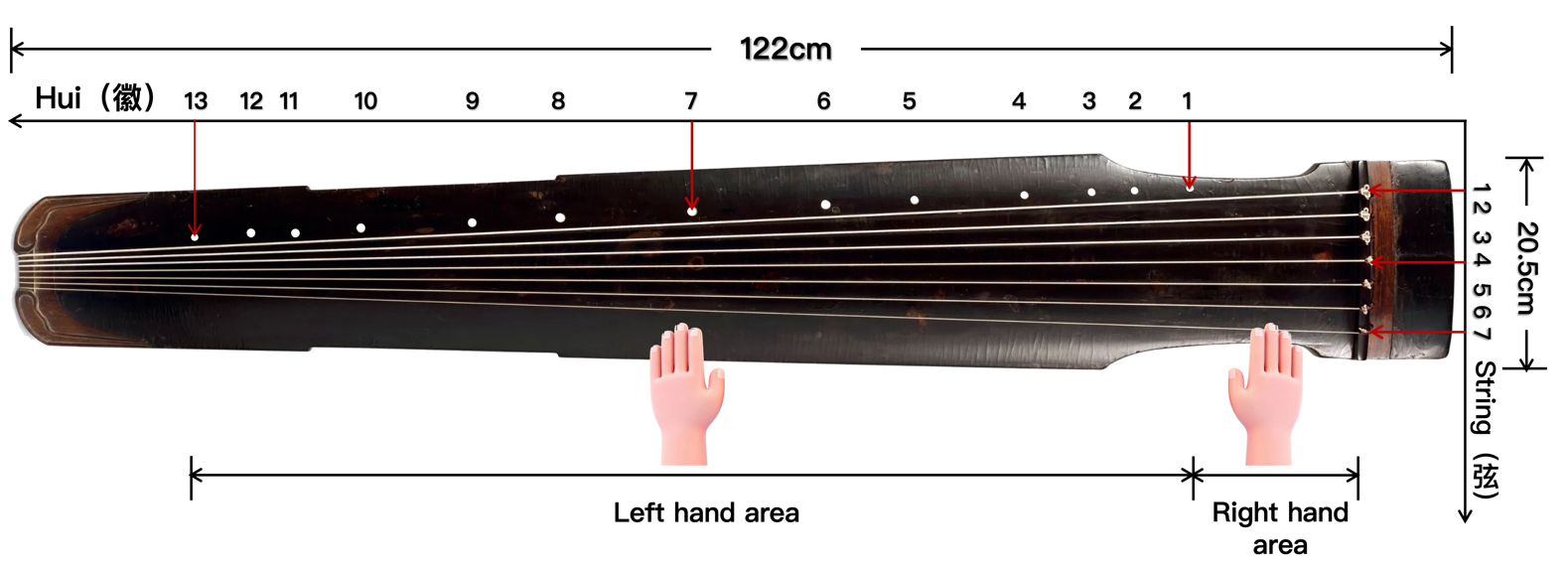}}
  \caption{The haptic robot finger is designed for guqin (a). The guqin playing robotic setting: The right hand is dexterous, with each finger featuring a downward-bending degree of freedom, and the thumb possesses a rotational degree of freedom. This configuration enables right-hand string plucking. The left UR5 robot is equipped with a haptic finger designed by our team to mimic the left-hand string-bending motion of a human. A calibrated {\color{red}Azure} Kinect V4 camera is positioned in the centre to capture time-synchronised visual and audio signals. (b). The guqin structure. }
  \label{fig:profile}
  \vspace{-0.5cm}
\end{figure}

\IEEEPARstart{T}{he} qin (Chinese: 琴), or guqin (古琴), 
a seven-string plucked zither (as shown in Fig. \ref{fig:profile}), is an ancient Chinese musical instrument with over 3000 years of history \cite{lixiangting}. Its performance art and style have been continuously refined over generations, profoundly embodying the Chinese national spirit, aesthetic sensibilities and characteristics of traditional art. In 2003, the guqin and its music were inscribed on UNESCO's Representative List of the Intangible Cultural Heritage of Humanity \cite{unesco}. Recently, Li \cite{Lihongliang} found that there are only 27 documents about guqin origin and development that can be found in the China National Knowledge Infrastructure database, raising concerns about the continuity of the tradition.


The guqin is primarily manifested as a solo art form. Its sound is not particularly loud; it is characterised by subtlety, elegance and restraint, seeking to express a poetic atmosphere that transcends mere sound. Its sound is simple and unadorned, yet resonant and lingering, imbued with endless meaning and resonance, and possesses immense evocative power.
Consequently, playing the guqin is technically demanding; the musician must demonstrate precise control of left-hand pressure, as well as mastery of sliding, vibrato, harmonics, whilst coordinating the flexible and rapid plucking of multiple strings with both hands.

To convey emotion during a performance, the musician must rely on the sensitive tactile perception of their fingertips to control the vibrations of the strings. When plucking with the right hand, they must at times employ the soft, delicate touch of the fingertips to produce a gentle, melodious sound, and at other times use the fingernails to produce a crisp, resonant tone. To evoke a particular atmosphere, when the left hand performs vibrato, it utilises the fingertips, fingernails, or half-tip-half-nail, creating a continuous, resonant effect through the friction generated as the hand glides along the body of the instrument.

Compared to the piano and other percussion instruments, guqin poses distinctive manipulation challenges: the left hand must precisely regulate pressing force, sliding motion, vibrato, and harmonics, while the right hand executes diverse plucking patterns. 
These demands make it a particularly suitable testbed for haptic sensing and dexterous robotic control.

In this paper, we present a design and validation of a biomimetic haptic fingertip, which is integrated into a proof-of-concept guqin system. 
Human fingertips combine slowly adapting pressure receptors and rapidly adapting vibration receptors to handle such tasks.
Inspired by this dual-channel sensing, we design a multimodal fingertip that integrates spatially resolved static pressure sensing with 
a broadband dynamic (vibration) sensing channel.
It not only incorporates tactile perception capabilities but also features elastic fingertips and anthropomorphic nail structures,
motivated by the distinctive ``half-finger, half-nail'' technique employed in guqin performance.
As shown in Fig. \ref{fig:profile}, we characterise the fingertip's tactile sensing performance and demonstrate its utility on a set of representative guqin string-contact
tasks: (i)~an open-string pluck comparison among three finger structures, evaluated with acoustic similarity metrics relative to a human reference; (ii)~a harmonic-tuning experiment; and (iii)~a tactile-event-triggered bimanual coordination experiment. We emphasise that this work validates a haptic fingertip on selected guqin contact tasks; it does not demonstrate complete guqin performance.

\section{Related Works}
\subsection{Haptic sensing for Dexterous Hands}
Compared to common two-finger grippers or simple end-effectors, multi-fingered anthropomorphic hands are more closely aligned with the human hand in terms of degrees of freedom, contact modes, and reachable motion space, and are in principle capable of executing fine-grained skills such as grasping, flipping, tapping, and plucking. At the same time, the high dimensionality and multiple contact points of dexterous hands significantly complicate modelling and control: the controller must handle strong coupling between joints, complex contact constraints, and tight temporal and force requirements imposed by the task. Thus, achieving human-level dexterity in robotic manipulation has long been a central goal in robotics.

For multi-fingered dexterous hands, recent studies have primarily focused on high-dimensional grasping and in-hand manipulation. Some works use reinforcement learning or optimisation-based methods to train dexterous hands in simulation for tasks such as object grasping, reorientation and rolling, and then transfer the learned policies to real hardware. Others incorporate tactile sensing and multi-modal feedback, and collect demonstration data via teleoperation to train end-to-end control policies. 

Recently, complementary work has investigated Sim2Real transfer for real-world piano playing \cite{robopianist2023}.
A flow-matching transformer policy trained on such datasets can yield agents that perform hundreds of pieces with notable generalisation to unseen songs \cite{dexterous_piano_scale}.
In \cite{SA}, Zhang et al. designed a rigid-soft dexterous hand dedicated to piano playing tasks. 
Wang et al. provided a learning-based method \cite{piano-tro} for a robot arm to play piano together with a human. 
Zeulner et al. train reinforcement learning policies in simulation and deploy them on a physical Allegro hand playing simple melodies such as “Twinkle Twinkle Little Star”, analysing how domain randomisation and dynamics modelling affect performance \cite{rl_real_piano}. 
Their results demonstrate that simulation-trained policies can produce meaningful real-world piano playing, while highlighting challenges in timing accuracy and robustness.
%

Published research on robotic chordophone performance has commonly implemented motor-based actuation in fingering mechanisms.
For instance, Park et al. introduced a robotic hand for a violin-playing robot, which features a finger with a three-axis load cell for precise force control \cite{violin-park}. 
Chaderfaux et al. analysed harpist forefinger movement and designed a robot finger for the harp plucking \cite{harp2012}. 
These kinds of dexterous hand designs utilise actuator-side torque sensing for joint force measurement and control, with limited integration of cutaneous tactile or skin-like sensors. %
For example, Yang et al. \cite{guitar} report an Expressive Robotic Guitarist whose fret-shifting and finger modules use torque control (and mode switching to position control) to implement pressing, damping, and release behaviours. 
Lin et al. combined the ideas of dual-arm teleoperation and visuotactile sensing for dual-arm manipulation skill learning in \cite{visuotactile_two_hands}.
However, relying solely on proprioception limits the robot's ability to perceive contact forces and their distribution, making it difficult to effectively transition the string state based on this most direct feedback—such as string-induced contact dynamics and acoustic–tactile coupling.
Görner et al. proposed a method for characterising the geometric models of Chinese Guzheng and their audio onset responses by employing robotic audio-tactile exploration with SynTouch BioTac fingertip sensors-equipped arm \cite{gorner2024pluck}. 
However, these works actually utilise only the multidimensional contact force information from fingertip tactile sensors. 
In the string-pressing scenario of the guqin, cutaneous tactile sensing must fulfil two core functions. 
First, it must accurately perceive the applied static force to ensure it remains within an optimal range—sufficient to alter the string's vibrating length without causing damage. 
Second, it must capture high-frequency vibration signals and transient contact disturbances in real time. 
This dynamic feedback is crucial for the fine adjustment of force and displacement during techniques like vibrato and glissando.

Our previous work has reported a multimodal biomimetic tactile fingertip \cite{hum2024,hum2025}, which demonstrated excellent performance in the task of tactile perception of liquid in containers by successfully predicting the liquid levels in various vessels. In contrast to traditional multimodal electronic skin devices based on printed thin-film electronics \cite{chun2021artificial} (Although they are highly skilled in the perception of textures), our highly integrated multimodal fingertip device has proven to be highly sensitive and robust, meeting the haptic perception requirements of humanoid robotics.
\begin{figure*}[thp]
\captionsetup[subfloat]{labelformat=empty}
  \centering
  \subfloat[\scriptsize{(a) Guqin Score "Jianzipu"}]{\includegraphics[width=\columnwidth]{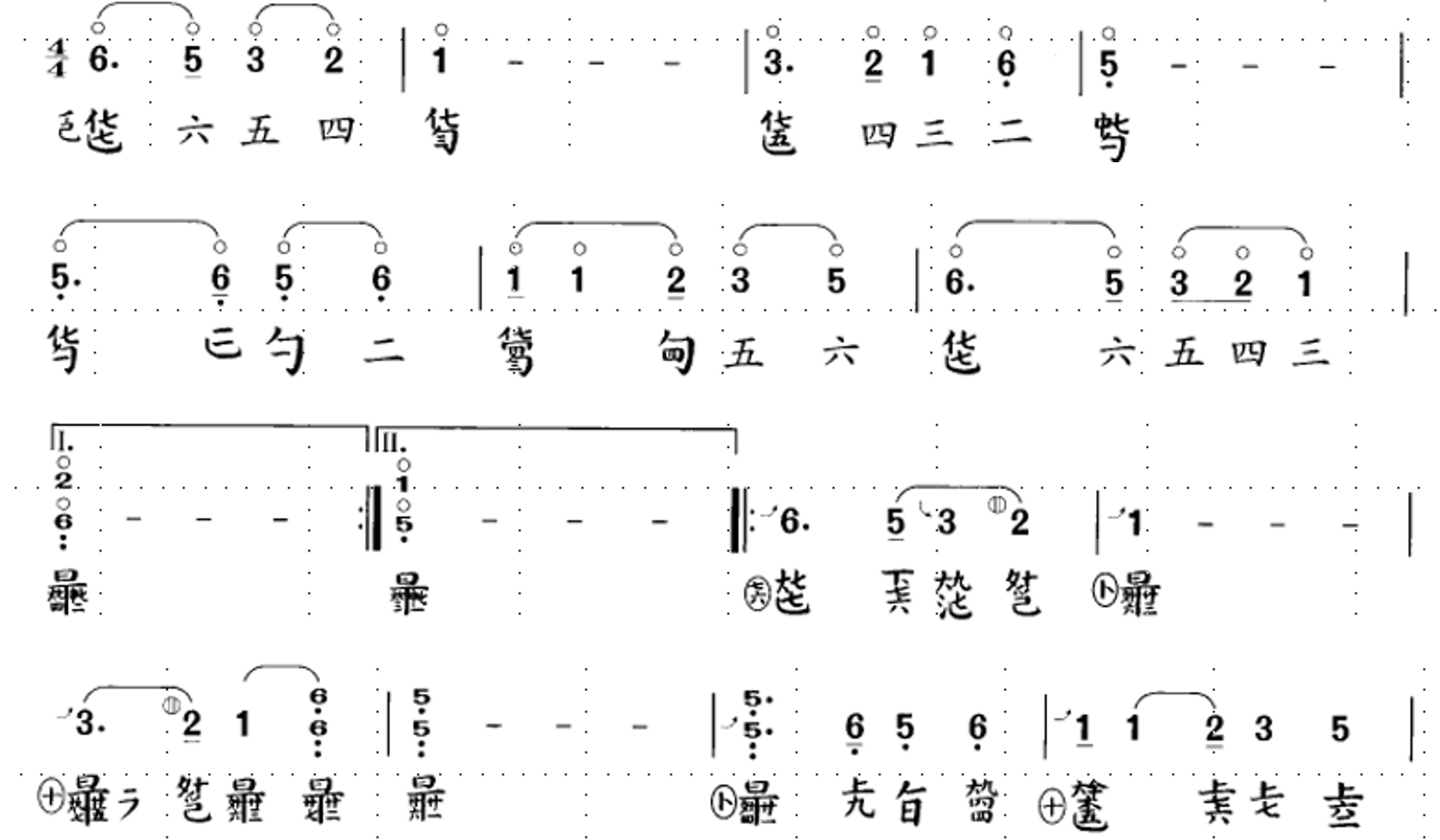}}
  \subfloat[\scriptsize{(b) Jianzipu Analysis}
  ]{\includegraphics[width=\columnwidth]{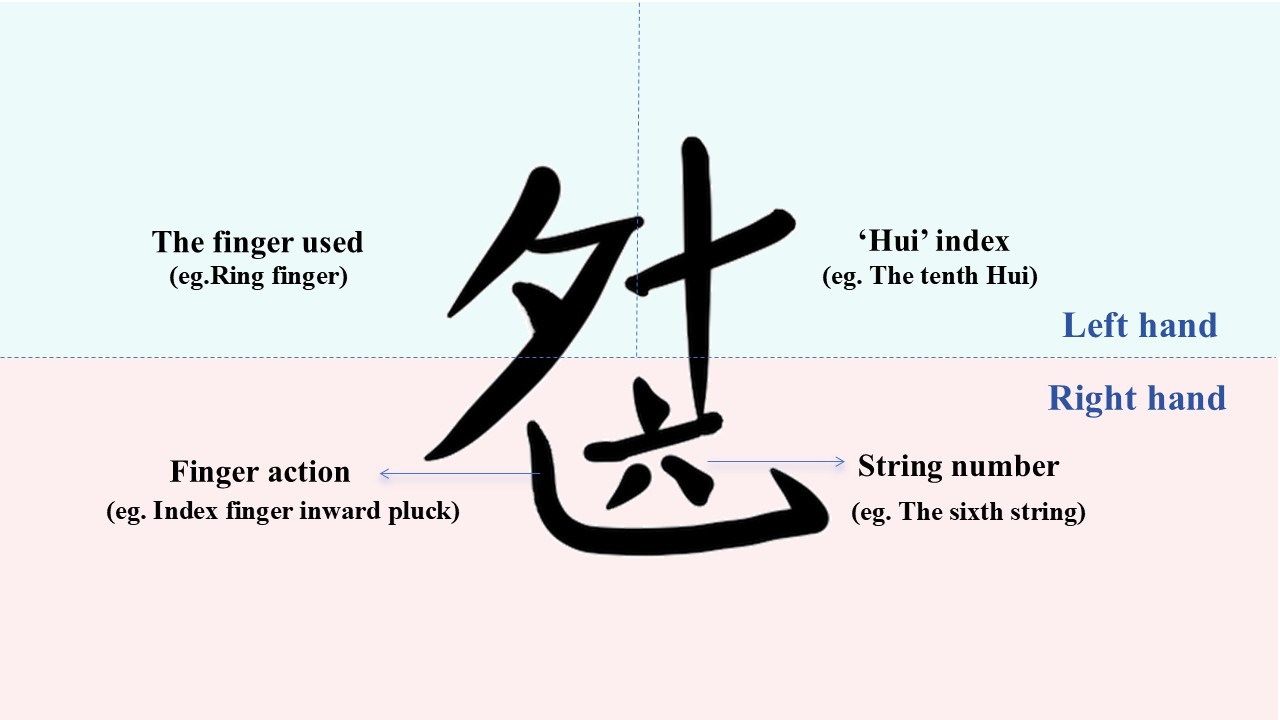}}
 \newline
  \subfloat[\scriptsize{(c) Guqin fingering}]{\includegraphics[width=\columnwidth]{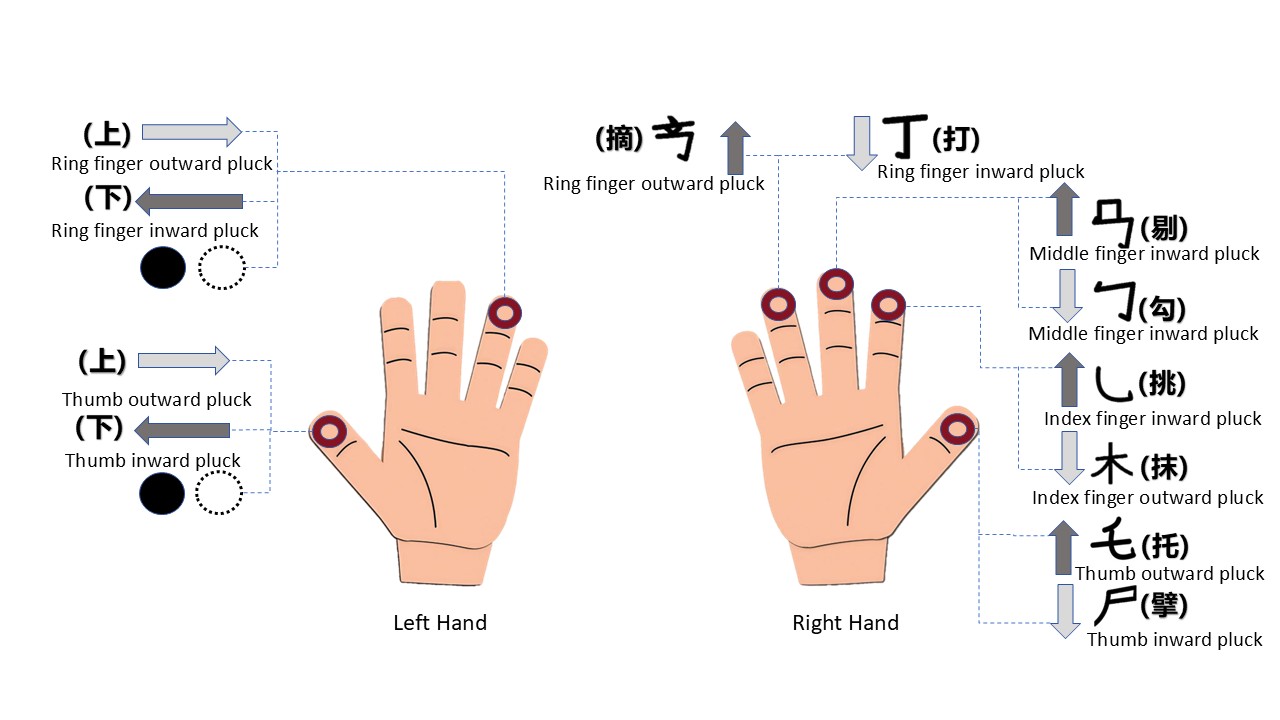}}
 \subfloat[\scriptsize{(d) Guqin Player's hand gestures are according to Jianzipu}]{\includegraphics[width=\columnwidth]{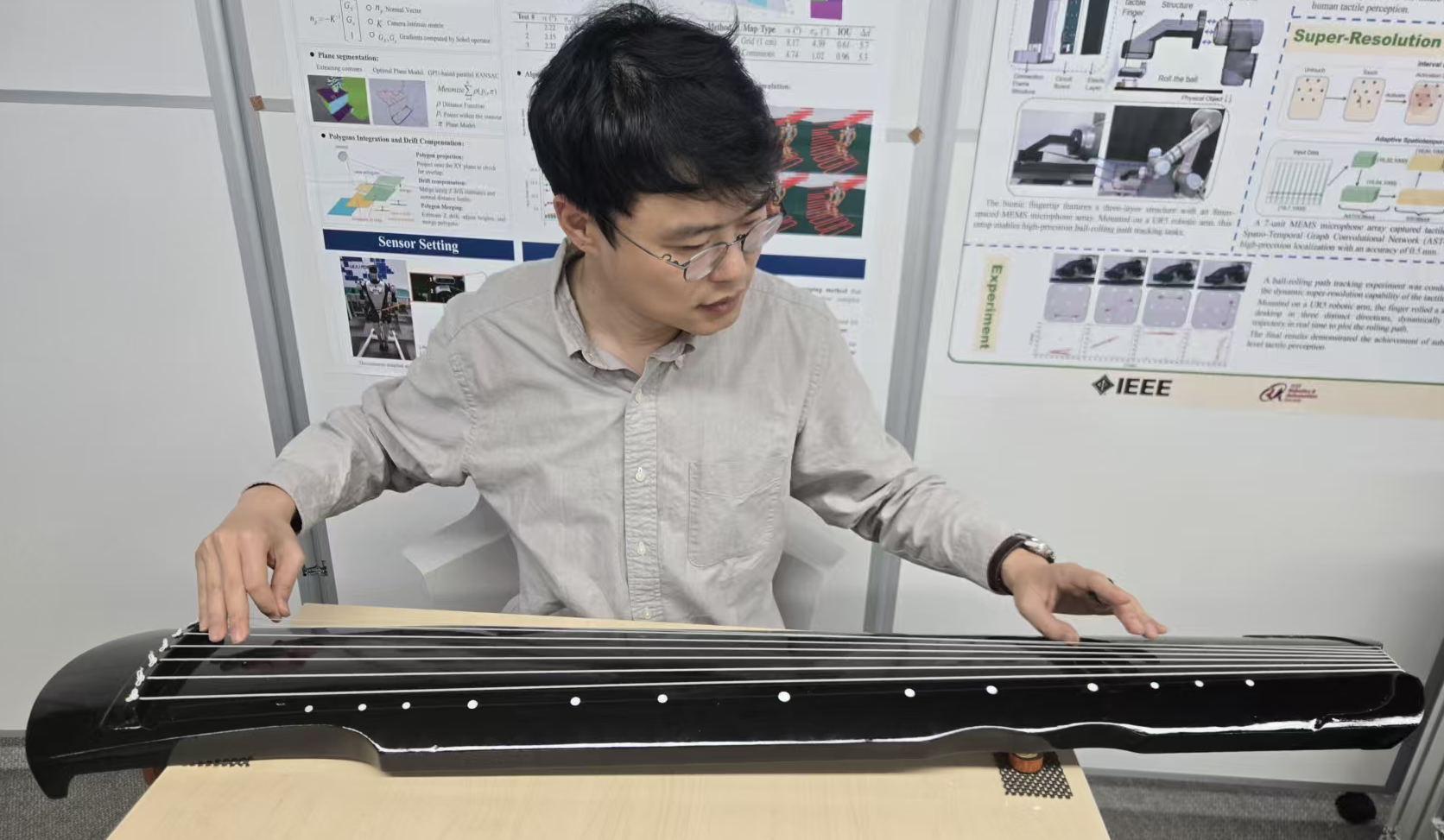}}
 \caption{The Jianzipu is a textbook documenting playing techniques, including the positions, sequence, and hand gestures for plucking the strings.}
 \label{fig:Jianzipu}
\end{figure*}
\subsection{Music-playing robots.}
Traditional industrial robots often rely on teach pendants or kinesthetic teaching: an operator directly guides the robot arm to a series of key poses, and the system records joint trajectories for later interpolation and replay. Such approaches work well in structured environments with repetitive tasks, but become expensive and inflexible when dealing with music playing. 
Because music instrument manipulation requires high-DoF systems and complex contact-rich tasks, considerable efforts have been expended across various domains, including mechanical structure, drive systems, skill acquisition, audiovisual perception, and tactile perception, to address the fundamentally distinct challenge of robotic instrument performance.
Perception capability is critical for music-playing robots.
Vision and audition are the most developed and most studied perceptual abilities endowed to robots \cite{my-survey-caai}. 

Musical performance is a representative scenario for such fine manipulation.
Many researchers have explored musical performance as a compelling demonstration of robotic manipulation capabilities.
In 2009, Otsuka et al. \cite{2009iros} presented a robot singer (HRP-2) using a beat tracking method to align polyphonic audio to the score. 
Görner et al. explored the string plucking motion control method \cite{guzheng} using a PR2 robot to pluck guzheng. 
Wang et al. \cite{guqin-toh} explored the virtual guqin playing method with 6-DoF haptic rendering of multi-region contacts between
a hand avatar and the strings. The authors of \cite{qin-hci} developed a human-computer-interaction system with digital twin tech for guqin education.  
Wakle et al. \cite{drummer} proposed a dielectric elastomer actuator for a robot drummer, which is promising to catch up with human drumming speed. 
Shahid et al. formulate drumming as sequential satisfaction of timed contacts, transform drum scores into “rhythmic contact chains”, and train a unified reinforcement learning policy to perform dozens of rock, metal and jazz pieces with high F1 scores \cite{robot_drummer}.

Taking these plucking instrument playing robot research as examples, a successful performance requires not only the fingers to strike keys or pluck strings accurately, but also to align precisely with the intended rhythm in time and to coordinate multiple fingers in space. If a \textbf{dexterous hand} can perform human-finger-like haptic sensing, its motion precision and control capability can be evaluated in some very intuitive real musical playing forms. 

\subsection{The robotic study of guqin and Jianzipu.}

After reviewing extensive literature on musical performance robots, we observed a significant focus on piano and percussion instruments. The guqin robot, much like the art of guqin itself, faces considerable challenges.
Numerous studies across musicology, fine arts, computer science \cite{qin-hci}, and the humanities and social sciences generally concur that the promotion and transmission of guqin art in contemporary society confronts major obstacles. 
These challenges primarily stem from the fact that the ``Jianzipu" records playing techniques, but does not provide detailed musical rhythms. This leads to the fact that guqin masters can only pass on their skills through one-to-one instruction. 
Although modern audio recordings are now available for reference, learners still find it difficult to compare them with Jianzipu and thus to teach themselves to perform guqin.    

See Fig. \ref{fig:profile} (b), guqin is approximately 122–125 cm in length, about 20 cm in width, and about 6 cm in thickness. It consists of the body and strings. The body is constructed by laminating a curved front panel and a flat back panel. 

Along the soundboard, thirteen circular markers (徽, \emph{hui}) are inlaid to indicate harmonic and stopping positions. The guqin is a fretless instrument: the \emph{hui} are inlaid position/harmonic markers. The instrument possesses seven strings, numbered from the side nearest the player inwards and from the lowest to the highest pitch: the first string to the seventh string.
One end of each string rests upon the bridge, secured by a threaded cord known as the ‘velvet cord’, which controls the pitch. This cord passes through the string hole on the soundboard and is fastened to the tuning peg. The other end of each string is looped around the string guide and wound around the tuning peg.
The material of the instrument body is the primary factor influencing its tonal quality. 
Typically, the soundboard is crafted from paulownia or fir wood, while the back and sides are made from catalpa or camphor wood. Traditional strings were made of silk; nowadays, people commonly adopt nylon-steel strings.

Guqin musical notation ``Jianzipu"  is the textbook documenting its playing techniques \cite{lixiangting}. As shown in Fig. \ref{fig:Jianzipu} (a) and (b), 
Jianzipu includes the positions, sequence, and hand gestures for plucking the string.
The abridged notation differs from both the Western staff notation and simplified notation, representing a fingering notation system passed down from ancient China to the present day. Each character records the plucking positions and fingering techniques for both hands on the guqin strings (see subfig (c)), thereby corresponding to the notes produced. As for the rhythm and melody of the entire piece, these often require the performer's own interpretation (as shown in subfig (d)).
Consequently, the guqin remains a relatively niche traditional instrument, its artistry relying on oral transmission from generation to generation of players.

Research exists on applying artificial intelligence to recognise musical notation, with most of the literature in Chinese and Japanese. In recent years, some scholars have attempted to employ artificial intelligence techniques such as image processing and pattern recognition to identify Jianzipu notation and automatically convert it into simplified notation, for example, \cite{Jianzipu-yolo} Jianzipu-YOLO and \cite{Jianzipu-cnn} Jianzipu-CNN.
However, no work has yet been done to realise the conversion of the action sequence from the simplified notation to dexterous hand manipulation and tactile feedback control. 
While prior robotic musical instruments have largely relied on position control and actuator-side torque sensing, our work focuses on cutaneous tactile sensing at the fingertip, aiming to capture both static force and fine-grained vibration cues during guqin string interaction.
Through performance demonstrations by the robot’s dexterous hands, people will be able to learn the guqin more conveniently and widely, thereby helping preserve guqin culture.

\section{Human-Like Haptic Finger Design}
The bottleneck in guqin robotics lies in achieving two distinct physical contact timbres: the stopped note
\footnote{We use ``stopped notes'' rather than ``fretted notes'' because the guqin is fretless.} and the harmonic.
Specifically, achieving precise string-pressing and pitch control in robotic instruments requires a haptic system that can simultaneously perceive static force and dynamic vibration while coordinating their complex relationship in real time. 

\begin{figure}[tbhp]
  \centering 
  \includegraphics[width=\columnwidth]{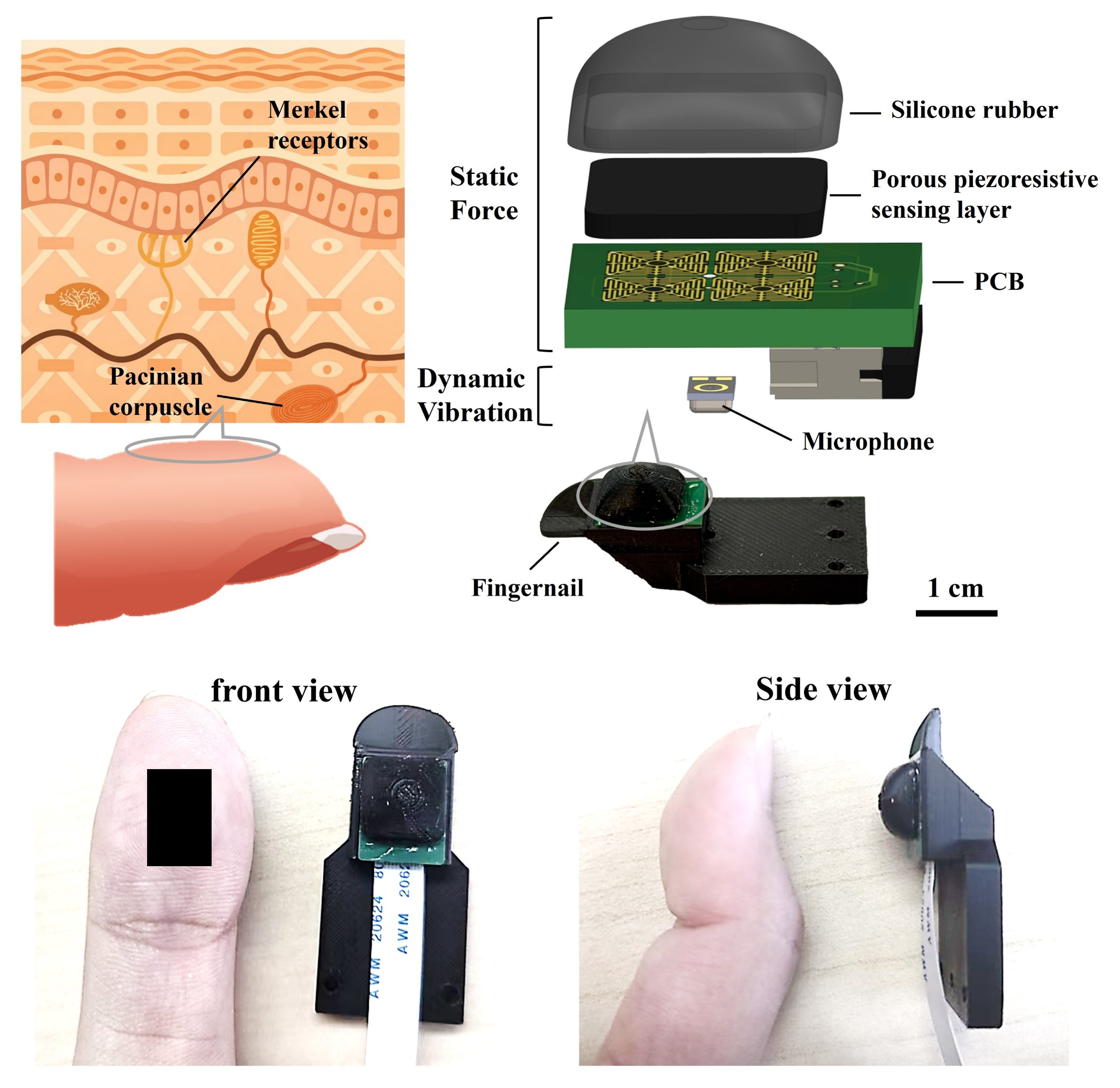}
  \caption{Schematic diagram of human skin illustrating mechanoreceptors: Merkel receptors for static force sensing and Pacinian corpuscles for dynamic vibration sensing. Exploded view of the multimodal biomimetic tactile fingertip, showing a layered assembly from top to bottom: a silicone rubber encapsulation shell, a porous piezoresistive sensing layer, and a rigid PCB with an integrated microphone, all mounted onto a rigid fingertip carrier with a biomimetic fingernail structure.}
  \label{fig:sensor}
\end{figure}
\subsection{Design Idea and Mechanism}
In traditional guqin performance, the right hand often uses longer nails for plucking, whereas the left hand typically maintains shorter nails to preserve fingerpad contact \cite{lixiangting}. 
Inspired by this observation, and as an engineering trade-off, we adopt a `half-nail, half-fingerpad' design: the nail-like part reduces friction and surface wear during sliding, while the soft part maintains stable pressing contact, which we found to yield more consistent tonal quality in our experiments. 
We do not claim that this structure directly replicates standard guqin left-hand technique; it is a biomimetically motivated design choice.

Thus, we developed a multimodal biomimetic tactile fingertip that combines spatially resolved static pressure mapping with a broadband dynamic sensing channel (validated over 1-100 Hz) mimicking the distinct functions of cutaneous slow- and fast-adapting mechanoreceptors. As shown in Fig. \ref{fig:sensor}, the tactile fingertip, measuring approximately 11 mm × 13 mm, is designed as a three-layer structure consisting of an encapsulating elastomer (Dragon Skin 10), a porous piezoresistive sensing layer, and an interdigitated electrode array patterned on a rigid PCB. All components are assembled within a rigid structural frame and bonded using silicone adhesive. The frame also incorporates a biomimetic fingernail structure. The porous piezoresistive sensing layer mimics the function of Merkel receptors in human skin by translating static contact forces into electrical signals via the piezoresistive effect. 

Under a low-pressure loading, the initial deformation can easily close the gaps between the interdigital electrode and porous piezoresistive sensing layer, resulting in the dropping of the contact resistance. Further increase of applied pressure starts to close the intrinsic hierarchical porous microstructure and continues to decrease the channel resistance of the pores. As pressure loading keeps increasing, based on the percolation effect, conductive particles are forced to pack more closely, enhancing the electrical conduction pathways, and the resistance of the sensing layer drops even more. To enable spatial force distribution sensing, the piezoresistive array was configured in a 4 × 4 layout, with each sensing unit measuring 2 mm × 2 mm.  Dynamic vibration sensing is achieved by integrating a capacitive silicon-based MEMS microphone (Goertek S15OB381) on the backside of the PCB, emulating the function of Pacinian corpuscles in human skin. 
This microphone has a nominal frequency response extending well beyond the audio band; however, our single-axis vibration platform permitted controlled excitation only over 1-100 Hz, and we therefore report the experimentally validated range as 1-100 Hz. This validated range is sufficient for the contact-detection and event-triggering role used in this paper, where the relevant cue is the transient onset of string contact rather than reconstruction of the full vibration spectrum.

\begin{figure}[tbhp]
  \centering 
  \includegraphics[width=\columnwidth]{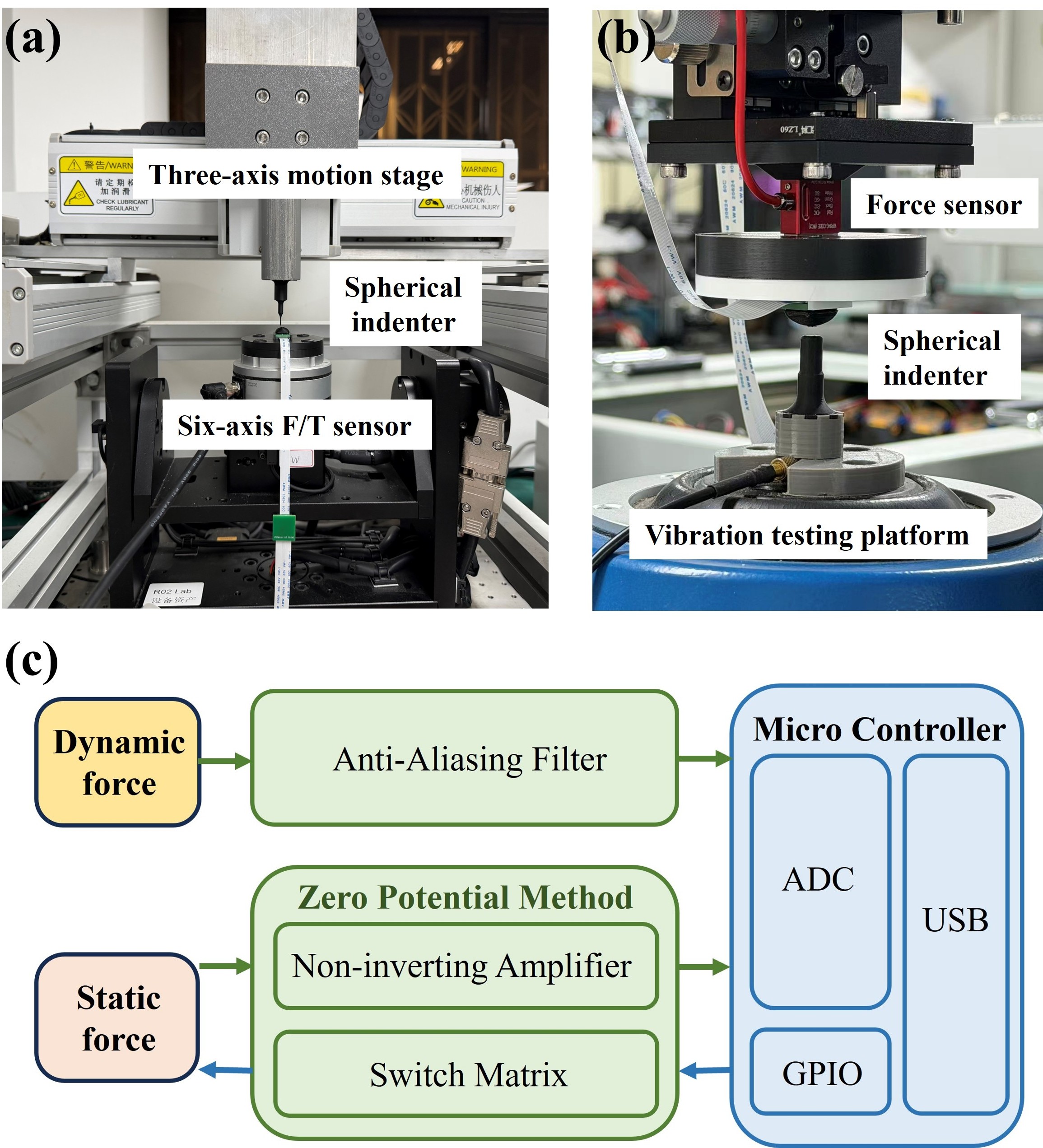}
  \caption{(a) Static force calibration platform. (b) Dynamic vibration testing platform. (c) Sensor data acquisition system.}
  \label{fig:biao}
\end{figure}

\subsection{Calibration of Tactile Fingertip}
The static calibration setup, shown in Fig. \ref{fig:biao} (a), consists of a three-axis motion stage, a six-axis force/torque (F/T) sensor (KUNWEI KWR75), and a spherical indenter. 
The motion stage provides micron-level, three-dimensional positioning accuracy, enabling precise alignment with the centres of all 16 units in the 4 × 4 electrode array. A spherical indenter (1.0 mm diameter) is mounted onto the motion platform via a connecting bracket. The six-axis force/torque (F/T) sensor is mounted on the base, with the biomimetic tactile fingertip sample positioned above it, to enable real-time measurement of the contact forces applied to the fingertip.

A single-axis vibration test platform with an excitation range of 1-100 Hz was employed to calibrate the dynamic response of the sensor (Fig. \ref{fig:biao} (b)). The data acquisition module synchronously records the multimodal outputs of the tactile fingertip at a unified sampling rate of 1000 Hz (Fig. \ref{fig:biao} (c)). The 16 piezoresistive channels are measured using a conventional zero-potential method circuit. 

\subsection{Force Sensing Performance Test}
\begin{figure*}[tbhp]
  \centering 
  \includegraphics[width=\textwidth]{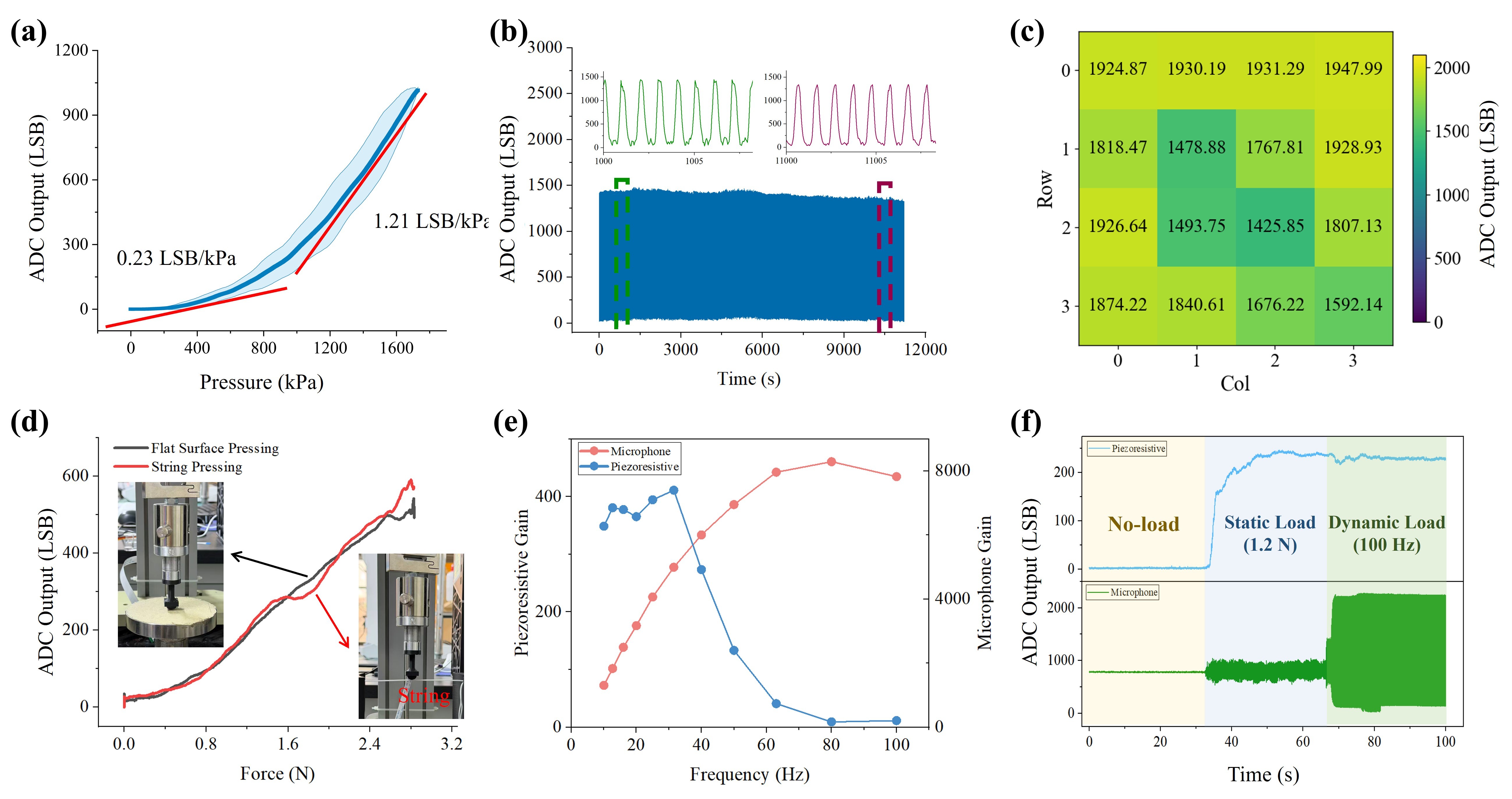}
  \caption{(a) Normalised average force–electrical response calibration curve for the 16 static sensing units. (b) Signal response of a single piezoresistive sensing unit over more than 10,000 pressing cycles. (c) Signal consistency test results across the 16 sensing units, with a 1.88 N force applied to each unit. (d) Comparison of static force sensor output signals under flat surface pressing and string pressing. (e) Gain distribution of the fast- and slow-adapting modal signals under external stimuli at different frequencies. (f) Signal characteristics of the fast- and slow-adapting modalities under static and dynamic loading. }
  \label{fig:ceshi3}
\end{figure*}

To quantitatively characterise the rich tactile information provided by the designed fingertip, we conducted a detailed evaluation of the sensing performance of both the fast- and slow-adapting modalities.

Fig. \ref{fig:ceshi3} (a) shows the normalised sensitivity ranges of the 16 sensing sites under applied force. Sensitivity was calculated from the slope of the pressure–ADC output response curve and is reported in LSB/kPa, where the ADC output was defined as the average signal across all taxels. The sensor exhibits a sensitivity of approximately 0.23 LSB/kPa in the 0-800 kPa range, which increases to 1.21 LSB/kPa in the 800-1600 kPa range. The durability of a randomly selected sensing site was evaluated through repeated contact testing for more than 10,000 cycles at 1 Hz (Fig. \ref{fig:ceshi3} (b)). Despite slight signal drift, the piezoresistive sensor maintained stable and reproducible performance, with fluctuations within 5\% and no signs of functional degradation. To verify signal consistency across the 16 sensing sites, a constant force was sequentially applied directly above each site, and the corresponding ADC output map is shown in Fig. \ref{fig:ceshi3} (c). The sensing array exhibited good overall consistency, while the slightly lower outputs at the central sites were attributed to the curved geometry of the silicone rubber, which was thicker at the centre and thinner toward the edges. 
Across this 16-site constant-force test, the centre coefficient of variation (CV) was 
$\approx${9.96}{\%}, the edge CV was $\approx${6.11}{\%}.
The baseline noise level was {3.178} {LSB} (RMS). Thus, the corresponding pressure resolutions are
$\approx$ {13.82} kPa (over 0-800 kPa) and $\approx$ 2.63 kPa (over 800-1600 kPa range).
This good consistency ensures the reliability of the sensor array and enables accurate perception of features from unstructured objects. 
Fig. \ref{fig:ceshi3} (d) compares the piezoresistive array responses during pressing against a rigid flat surface and a suspended flexible string. The close agreement between the two curves indicates that the fabricated fingertip device can reliably operate in flexible interaction scenarios such as string pressing, while also confirming the validity of the previously established calibration. 

To characterise the frequency responses of the fast- and slow-adapting modalities and their transient dynamics, a single-axis vibration test platform with an excitation range of 1–100 Hz was employed. As shown in Fig. \ref{fig:ceshi3}(e), the gain of the piezoresistive signal decreases significantly at frequencies above 30 Hz, whereas the microphone signal is enhanced over the same range. These results indicate that the two modalities exhibit complementary frequency-response characteristics with minimal mutual coupling. Fig. \ref{fig:ceshi3}(f) further illustrates the signal characteristics of the fast- and slow-adapting modalities under combined static and dynamic loading, confirming their coordinated decoupled responses. Specifically, increasing the vibration frequency does not affect the static piezoresistive response, while the applied static load has a negligible influence on the dynamic microphone output. The small residual fluctuations are mainly attributed to noise from the vibration platform.
\subsection{String Pressing Test}
\begin{figure}[tbhp]
  \centering 
  \includegraphics[width=\columnwidth]{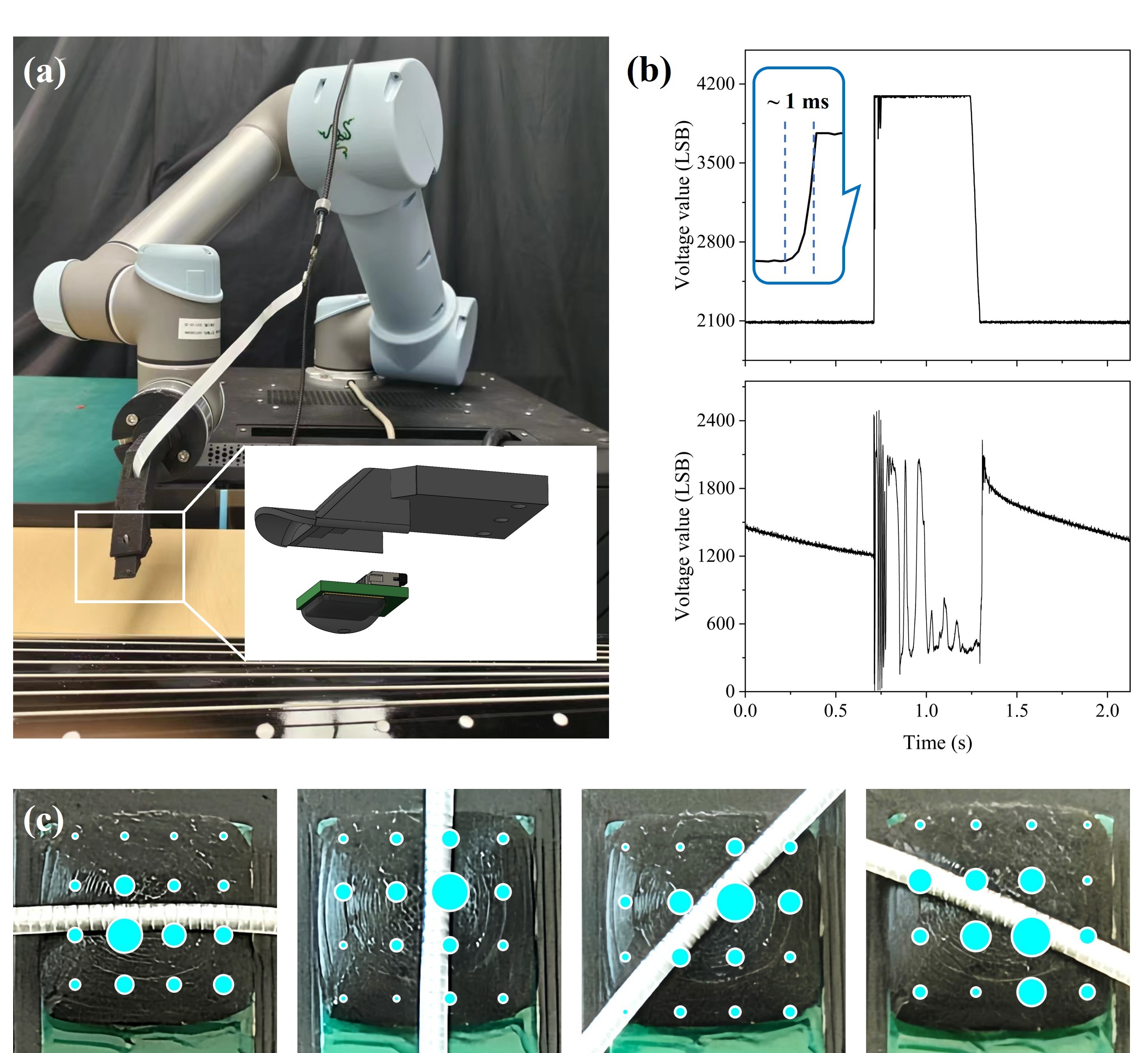}
  \caption{(a) Schematic of the robotic arm pressing a string. (b) Timing diagram of static and dynamic forces during string pressing. (c) Spatial distribution of static pressure during string pressing at different contact angles. }
  \label{fig:mapp}
\end{figure}
The multimodal biomimetic tactile fingertip was mounted on the end effector of a UR5 robotic arm via a 3D-printed adapter before task execution (Fig. \ref{fig:mapp} (a)). Practical pressing tests demonstrated that the tactile sensor can simultaneously capture both static and dynamic pressure stimuli. The dynamic outputs appear as pulsed signals, analogous to biological responses associated with contact onset and offset, while also encoding information about the frequency range of string vibrations (Fig. \ref{fig:mapp} (b)). The static results further show that the tactile fingertip can reliably capture the loading and unloading of static force. Moreover, the static force map obtained during contact with a string (Fig. \ref{fig:mapp} (c)) provides information such as the relative finger–string position and the applied force magnitude, thereby supporting more dexterous manipulation in musical performance tasks.

\section{ Experiment and Evaluation}
The guqin's range spans from C to D2. Beyond its seven open strings, it possesses 91 harmonics and over a hundred stopped notes.
Designing tactile-based dexterous fingers enables the robot to produce these three types of guqin tones (open strings, stopped notes, and harmonics).
To demonstrate the capabilities of the designed biomimetic haptic finger, we integrated it into a proof-of-concept guqin system as shown in Fig. \ref{fig:profile}.  
In our experimental setting, the right hand is a BrainCo Revo1 dexterous hand connected with an elfin3 manipulator. 
The left hand is the proposed haptic finger mounted on a UR5 manipulator. 
An Azure Kinect V4 RGB-D camera is mounted above the guqin to capture images of the strings whilst simultaneously recording audio signals. 
This camera can collect visual and auditory information in real time, which is then synchronised with tactile signals from the proposed finger, thereby enabling the fusion of these three sensory modalities.
The entire system software is integrated into ROS2, including the motion planner, controller, and multi-sensor perception.

\subsection{Comparison of Finger Structures}
\begin{figure*}[tbhp]
  \centering
  \includegraphics[width=\textwidth]
  {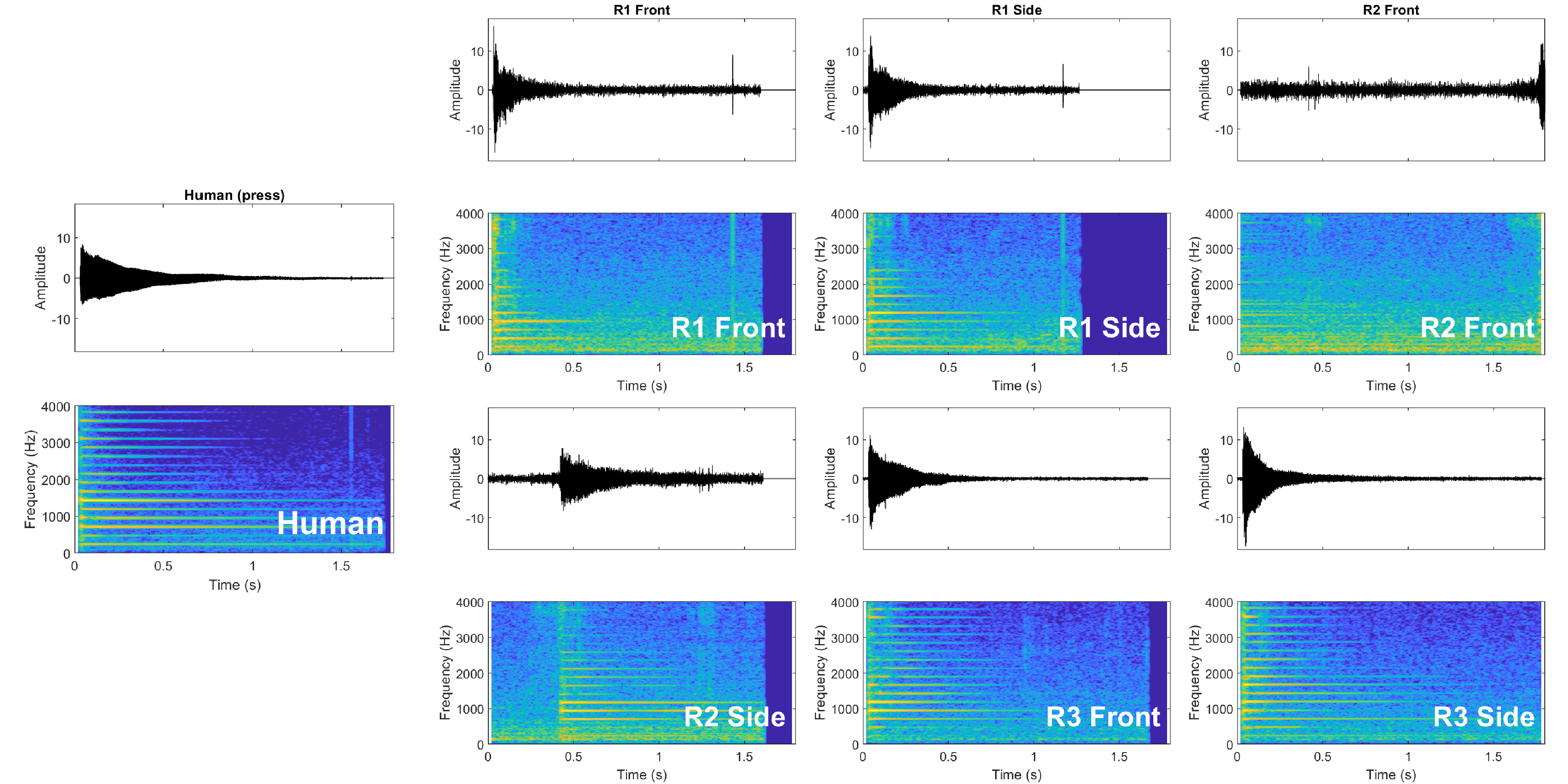}
  \caption{Waveform--spectrogram comparison of human open strings versus three robotic fingertips under front/side contact configurations (onset-aligned; fixed dynamic range).  Plots are representative single trials.}
  \label{fig:fig_29}
\end{figure*}

\begin{figure}[tbh]
  \centering
  \includegraphics[width=\columnwidth]{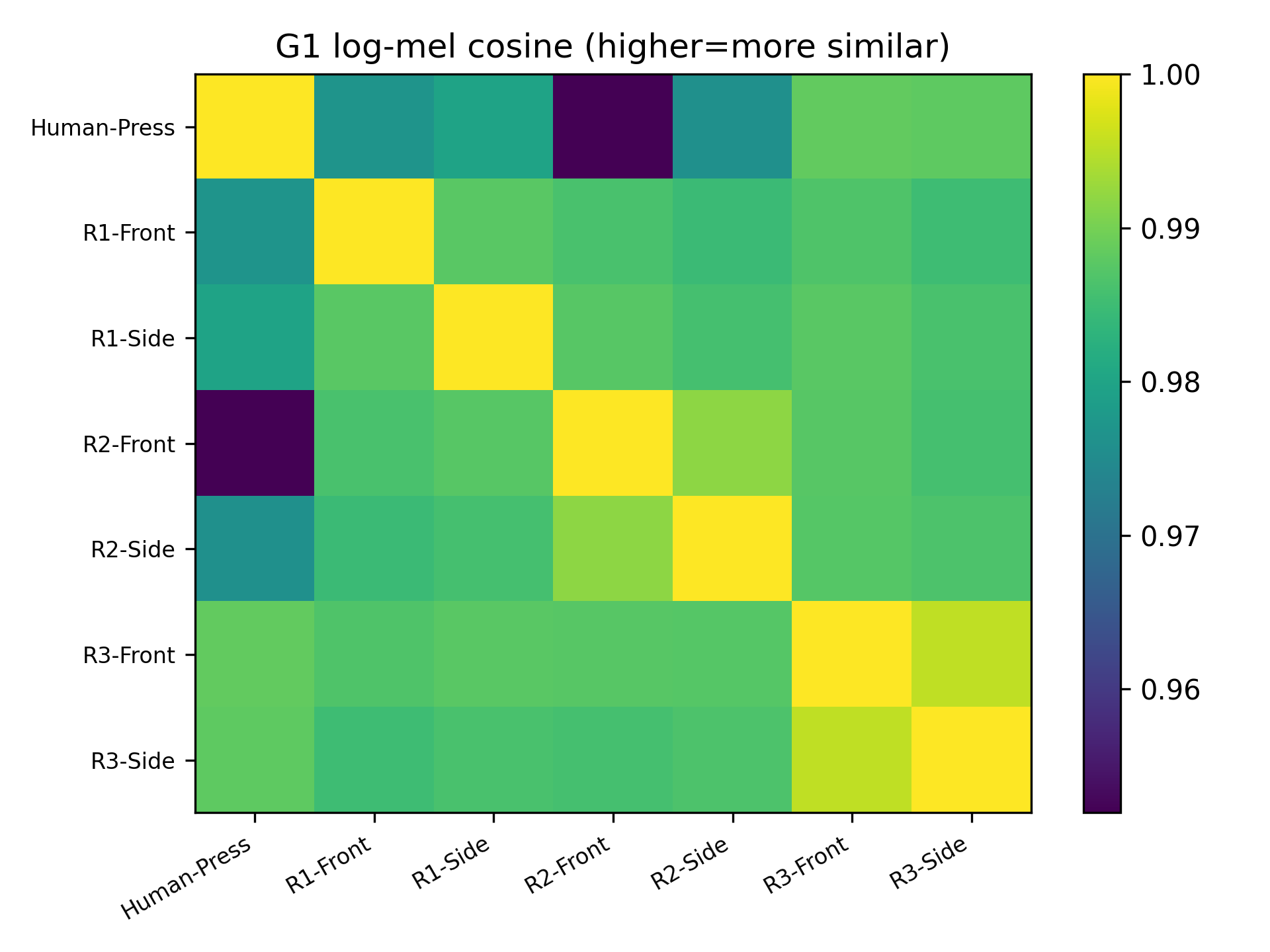}
    \caption{The Similarity heatmap for the G1(1st \emph{hui}, 4th string) open string pluck experiment(higher indicates greater spectral similarity).}
  \label{fig:fig_26}
\end{figure}

To design tactile fingertips better suited for guqin performance, we experimented with three fingerpad designs as
R1: a 6 $\times$ 6 resolution flat surface, R2: a 4 $\times$ 4 flat surface with a nail, and R3: a 4 $\times$ 4 curved surface with a nail.
To validate the proposed fingertip on guqin string-contact tasks, we used the identical robotic system configuration to test the three finger types.
The three finger structures (R1, R2, R3) were qualitatively exercised across the G1(1st-\emph{hui}, 4th string) open-string pluck experiment. 
Each open string trial was compared against live human performance, yielding the following quantitative acoustic comparison reported in  Table~\ref{tab:g1_press_pluck_full} and Figs.~\ref{fig:fig_29}--\ref{fig:fig_26}. 
The stopped-note and harmonic results are presented separately in Sections~\ref{sec:coord} and~\ref{sec:harmonic}. 

\begin{table*}[tbhp]
\centering
\caption{Open-string pluck comparison (reference: Human-Press). Higher log-mel cosine and envelope correlation indicate greater similarity; lower DTW chroma cost indicates greater melodic similarity.  Values are means over $10$ trials.}
\label{tab:g1_press_pluck_full}
\scriptsize
\setlength{\tabcolsep}{6pt}
\renewcommand{\arraystretch}{1.12}
\begin{tabular}{lrrrrrrrr}
\hline
Target ID & LUFS & Centroid(Hz) & log-mel $\uparrow$ & DTW $\downarrow$ & EnvCorr $\uparrow$ & Shift(s) & $\Delta$LUFS & $\Delta$Centroid \\
\hline
R3-Front & -29.384 & 821.049 & 0.988 & 0.096 & 0.937 & 0.511 & -4.014 & -49.969 \\
R3-Side  & -30.029 & 829.295 & 0.988 & 0.087 & 0.869 & 0.494 & -4.659 & -41.724 \\
R1-Side  & -40.476 & 834.280 & 0.980 & 0.162 & 0.860 & 1.434 & -15.106 & -36.739 \\
R1-Front & -43.848 & 670.265 & 0.977 & 0.104 & 0.642 & 0.854 & -18.478 & -200.754 \\
R2-Side  & -44.018 & 655.749 & 0.976 & 0.100 & 0.646 & 2.731 & -18.648 & -215.270 \\
R2-Front & -45.603 & 638.449 & 0.952 & 0.323 & 0.543 & -0.231 & -20.233 & -232.570 \\
\hline
\end{tabular}
\end{table*}

\begin{figure}[tbhp]
  \centering 
  \includegraphics[width=\columnwidth]{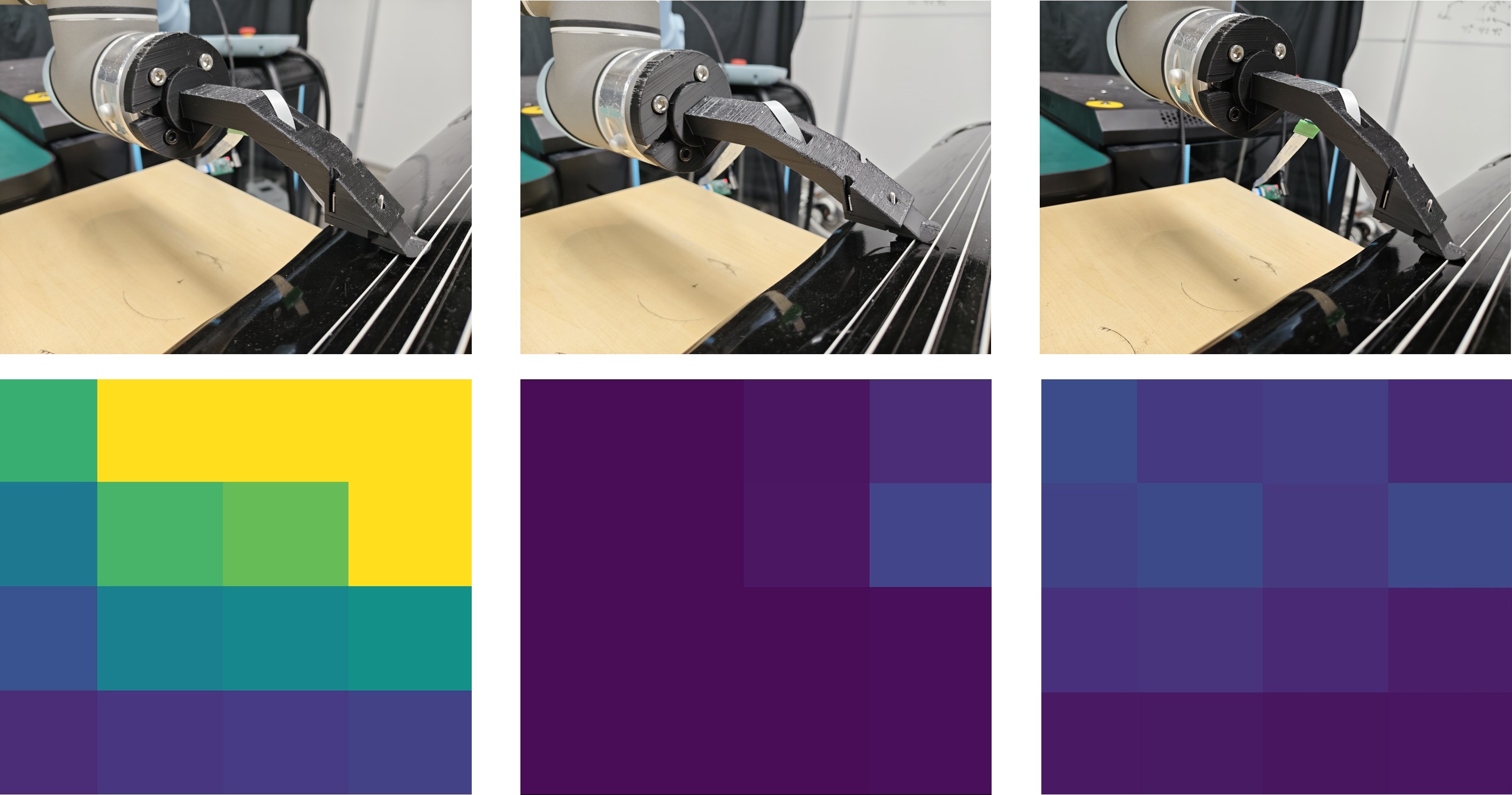}
  \caption{The Haptic sensing feedback in guqin playing. 
  Here are three types of string pressing skills that the robot mimics human skills, from left to right: fingertip, half-finger half-nail, and full nail.}
  \label{fig:hap1}
\end{figure}

Firstly, we observed that the flat fingertip design yields clearer tactile feedback (due to greater contact with the instrument surface), but this also results in an excessively high coefficient of friction (between raw lacquer and silicone). Whilst pressing down the string and sliding it left or right,  this readily causes damage to the silicone coating on the sensor's surface. Thus, the R3 curved fingerboard design offers superior performance during pressing and vibrato techniques.
Fig.~\ref{fig:fig_29} compares the human playing baseline with robotic trials using these 3 fingertips under front/side contact configurations. The aligned waveforms highlight differences in attack timing and decay behaviour, while the spectrograms reveal changes in harmonic structure and broadband noise across conditions.
The waveforms and spectrograms in Fig.~\ref{fig:fig_29} are representative single trials (onset-aligned, fixed dynamic range), whereas the similarity metrics in Table~\ref{tab:g1_press_pluck_full} and the heatmap in Fig.~\ref{fig:fig_26} are the mean across the $10$ trials.

\begin{figure*}[tbh]
  \centering
  \includegraphics[width=2\columnwidth]{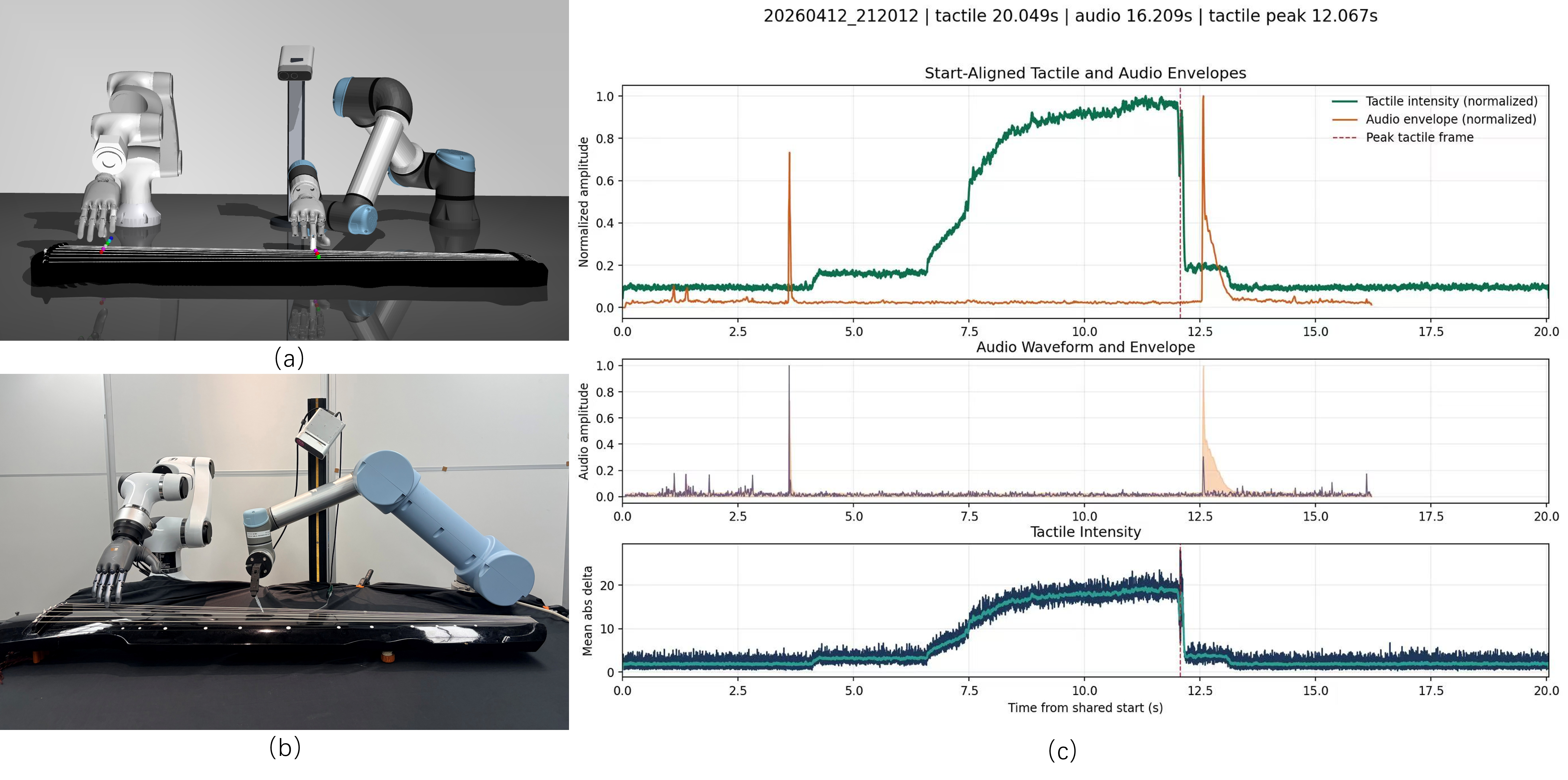}
    \caption{(a) is the system simulation, (b) shows the real robot system. (c) A Tactile-Event-Triggered Bimanual Coordination Experiment.}
    \label{fig:robot-taticle}
\end{figure*}

Secondly, the bigger design (R1) hinders finger movement between densely packed strings (as the strings are more tightly spaced on the left side).
As shown in Fig.~\ref{fig:fig_26}, the R3 variants
are closer to the human reference than the other conditions, while R2-Front is the most distant. R3-Front and R3-Side cluster tightly together, indicating that their similarity is driven by consistent spectral shape rather than by timing alignment.

Thirdly, the nail design lends the robot a more lifelike performance effect.
See Fig. \ref{fig:hap1}; we compared the three string pressing actions using the R3 finger. These actions are mimicked by human player motion: front press (full fleshy part of finger), side press (half nail, half fingertip) and tip fingertip press. 
Referring to the results in Fig. \ref{fig:fig_26}, 
among the tested configurations, R3-Side yielded the acoustic features closest to the human reference. Half the nail's contact reduces friction with the guqin board,
while half the fingertip's contact presses the strings more firmly against
the board, producing a smooth tone with the smallest acoustic deviation
from the human reference in our comparison.

The open string comparisons are listed in Table~\ref{tab:g1_press_pluck_full} 
across loudness, spectral centroid, and the similarity scores. 
R3-Side is consistently closest to the human reference across all reported metrics,
while R2-Front is consistently the most distant.

\subsection{Tactile-Event-Triggered Bimanual Coordination}
\label{sec:coord}

This makes harmonic playing a suitable scenario for demonstrating tactile-event-triggered bimanual coordination using the designed finger.
The control is event-triggered: we compute the aggregated left-hand tactile intensity (summed across the 16 taxels) and monitor its band-limited rate of change. When this rate exceeds a fixed threshold---denoting contact onset--the right hand is commanded to execute a single pre-planned pluck. We emphasise that the tactile signal triggers only the \emph{timing} of the next action; it does not continuously adjust pressing force, finger position, or release timing.
Accordingly, we describe this as tactile-event-triggered bimanual coordination rather than continuous tactile force/trajectory (servo) control.

As shown in Fig. \ref{fig:robot-taticle}, (a) is the system simulation, (b) shows the real robot system.
The experiment was set up such that the robot’s left-hand tactile fingertip moved to press a guqin string, generating a tactile signal. Upon receiving this signal, the right arm executed a movement, and the middle finger of the right hand plucked the string downwards, producing a sound wave. 
 The experiment was repeated 10 times.

The mean delay between the detected tactile contact onset and the audio peak was
{481} $\pm$ {56} {ms} across the 10 trials, measured from the time-synchronised tactile and audio streams, see figure (c).
These results indicate that the event-triggered bimanual coordination is repeatable with low timing variability across trials, and that the designed fingertip provides a reliable contact-onset signal for triggering the right-hand action.

\subsection{Harmonic performance test}
\label{sec:harmonic}

\begin{figure*}[tbhp]
  \centering
  \includegraphics[width=1.8\columnwidth]{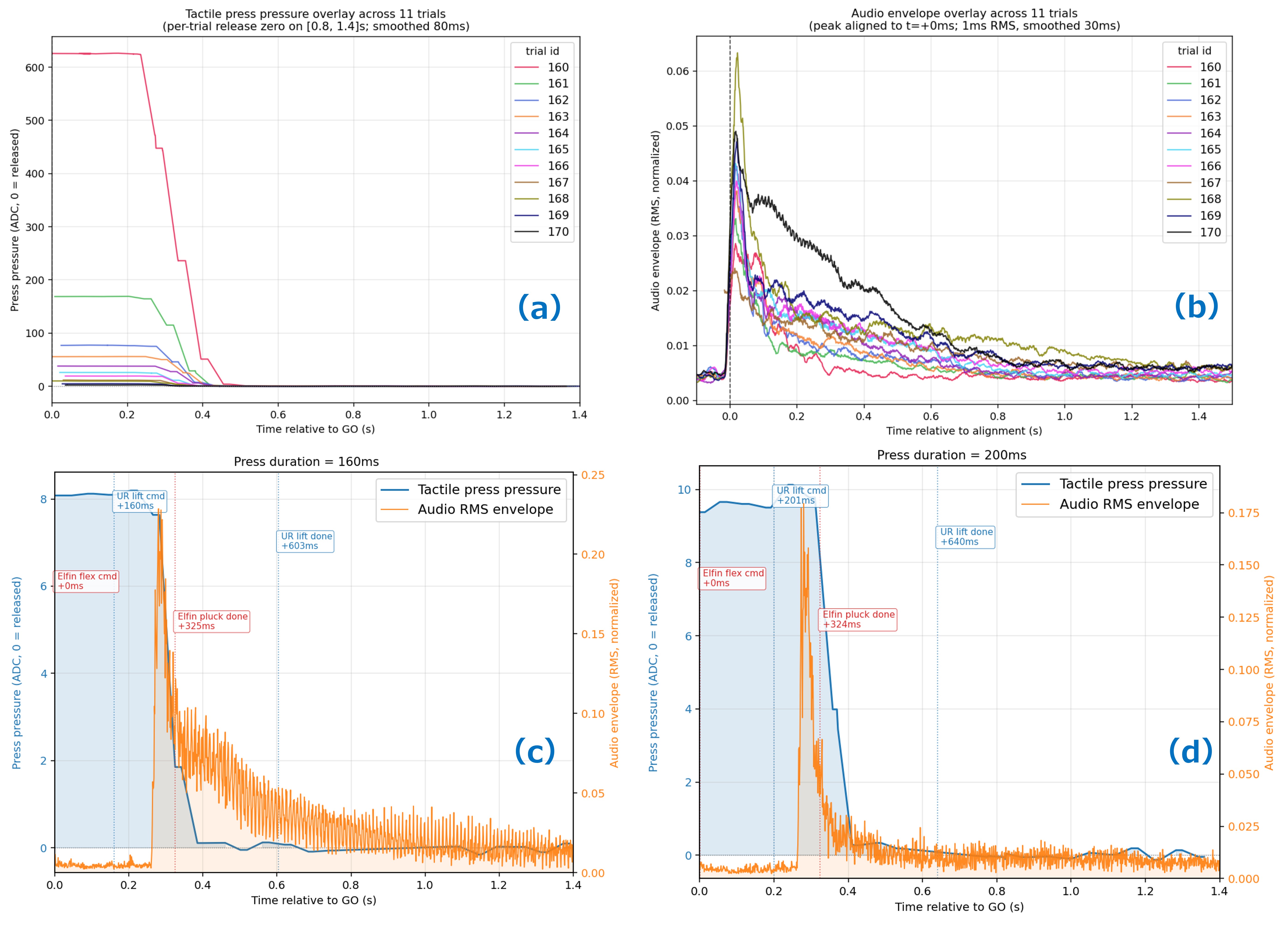}
    \caption{Harmonic playing parameter tuning experiments. (a) and (b) show 11 trials of different sound waveforms and haptic curves corresponding to the tactile sensor pressing the string down from 0 to 10 mm within a 1.5 sec time window. (c) and (d) show the Waveforms of the harmonics produced by the right hand 160 ms and 200 ms after the left hand has pressed the string. }
  \label{fig:harmonic}
\end{figure*}

Furthermore, to produce harmonics, the left hand must lightly press the string; after the right hand plucks it, the left hand should pause briefly before releasing the string, allowing the vibration to be released by the resistance of the fingertip, thereby producing a delicate and ethereal tone. 

To assess the effect of varying finger pressure on harmonic performance, we tested harmonic production at string depths ranging from 0-10 mm while increasing the delay of the right-hand pluck. 
The results of 11 trials at different depths are shown in Fig. \ref{fig:harmonic} (a) and (b): At 0 mm, the sound is completely open; as the pressure depth increases, the harmonic tone peaks at 8 mm, then gradually diminishes until, at 10 mm, the string touches the guqin wood board, and the sound transitions to a stopped note.
As the fingers of the right hand lack tactile sensation, we determined the time difference by adjusting the timing of the right-hand pluck and immediately lifting the left hand; this time difference gradually increased from 120 to 200 ms. 

To help select the left-hand release-delay parameter, we conducted a preliminary perceptual check. 10 volunteers participated. 
None had formal guqin training; five of them reported general musical experience.
Each participant first heard a reference recording of a professional guqin
harmonic, then listened to robot renditions produced at the candidate release delays (5 samples of 120, 140, 160, 180 and 200 ms delays) and selected the rendition judged most similar to the reference; the samples are played in random order.
Finally, 8 of the 10 participants selected the 160 ms condition as most similar to the reference.
These results show that the harmonics were judged perceptually closest to the reference among the tested delays when the left hand was lifted at 160 ms. Because the panel did not include expert guqin performers and the procedure was not a formal perceptual study, we report this result as preliminary evidence supporting parameter selection, rather than as validation of human-level or authentic guqin performance.

In addition, we show the tactile sensing and audio envelope data of 160 and 200 ms in Fig. \ref{fig:harmonic} (c) and (d). According to these sound data observed in the experiment, we find that: reducing the delay in lifting the left hand results in a higher-pitched sound and a louder tone, which sounds like an open string; conversely, extending this delay causes the audio waveform to drop in pitch, and the sound resembles a deep, muted stopped note.


\section{Limitation and Discussion}
There are many existing limitations for the guqin-playing robotic system. 
Firstly, current robotic dexterous hands are subject to significant hardware limitations; the dexterous hand used in this paper is not capable of performing flexible and high-speed plucking motions.
For example, current robot hands cannot perform the advanced `wheel fingering' technique used in the famous guqin piece `Flowing Water'. This technique involves rotating the wrist so that the index, middle and ring fingers move in succession across multiple strings, much like three points on a spinning wheel. To replicate this movement, the robot wrist joint has to function as a spherical omnidirectional joint. Such spherical joints are extremely rare in current robotics applications, with most remaining at the research and validation stage in mechanical design \cite{tadakuma-tro}.
However, the Jianzipu records the sequence of movements involved in guqin playing; it is precisely through these rapid sequences of movements that the performer conveys a portion of the emotions they wish to express, and this emotional expression constitutes the artistic essence and distinctive feature of guqin music.
The mechanical constraints on these robots’ wrists and fingers prevent them from simply replicating the movement trajectories of human performers,
which presents us with a significant challenge: planning the robot’s performance motions in accordance with its kinematic constraints, whilst ensuring they not only adhere to the rhythm of the piece but also attempt to convey the human emotions expressed in the music, thereby preserving the artistic essence of classical music.

Secondly, existing sensor fusion systems remain imperfect. High-level guqin performance requires extremely rapid hand movements, complex fingering techniques, and frequent, precise physical contact. The 30 Hz visual input limits our visual-tactile-auditory fusion output, meaning that even with a detailed guqin simulation model, it is difficult to achieve precise, coordinated bimanual control of playing movements at 100–200 Hz.

Thirdly, guqin playing evaluation is an open problem for robotic systems.
Compared to the existing piano-playing robot research, such as \cite{pianoJudges, trans,pianoprecision}, 
it is difficult to make quantitative comparisons and assessments of guqin performances.
This is because, firstly, the sound of the guqin originates from the vibration of the entire wooden body; as each guqin instrument is handcrafted, with variations in materials and dimensions, the tonal characteristics of each guqin differ to some extent.
Furthermore, the sound produced by the guqin is continuously evolving, unlike that of the piano, where each note is discrete; this makes it more difficult for AI and robotic systems to analyse and evaluate it, which poses a number of challenges for robot learning methods.

The proposed multimodal haptic fingertip may also be useful beyond musical
performance---for example, in tactile-rich medical and assistive tasks
such as palpation-based examination and massage, where human-like contact
sensing and learned manipulation skills are required. We leave such
applications to future work, once sufficient embodied haptic data has been
collected.

\section{Conclusion and future work}
In this paper, we have designed and validated a haptic finger with a humanoid structure.
It features a soft fingertip and nail structure, which we validated using a dual-arm robot guqin playing system. 
Our experiments demonstrate that the proposed haptic fingertip reliably executes selected guqin string-contact tasks---open strings, harmonics, and stopped notes---in our setup. We note that a complete guqin performance (e.g., fast multi-string techniques and expressive interpretation) is beyond the present scope, as discussed in the Limitations.

Future work includes developing a Vision-Language-Action module for robot guqin piece playing. 
Using the VLA models, align the audio of the guqin performance to the corresponding Jianzipu, then align this to the corresponding video of a human playing the instrument, whilst simultaneously identifying the contact between the human hand and the corresponding strings. Through a tri-modal (audio-vision-Jianzipu) aligned fingering analysis, the hand motion sequences can be transferred to robot guqin playing motion trajectories.

\section{Acknowledgment}
We would like to thank Zhen Wang and Xinyue Wei for their assistance during the experiment. We would also like to express our sincere gratitude to the reviewers and editors of ToH; your encouragement and suggestions were invaluable in improving this study. 

\bibliographystyle{IEEEtran}
\bibliography{ref}
\begin{IEEEbiography}[{\includegraphics[width=1in,
clip]{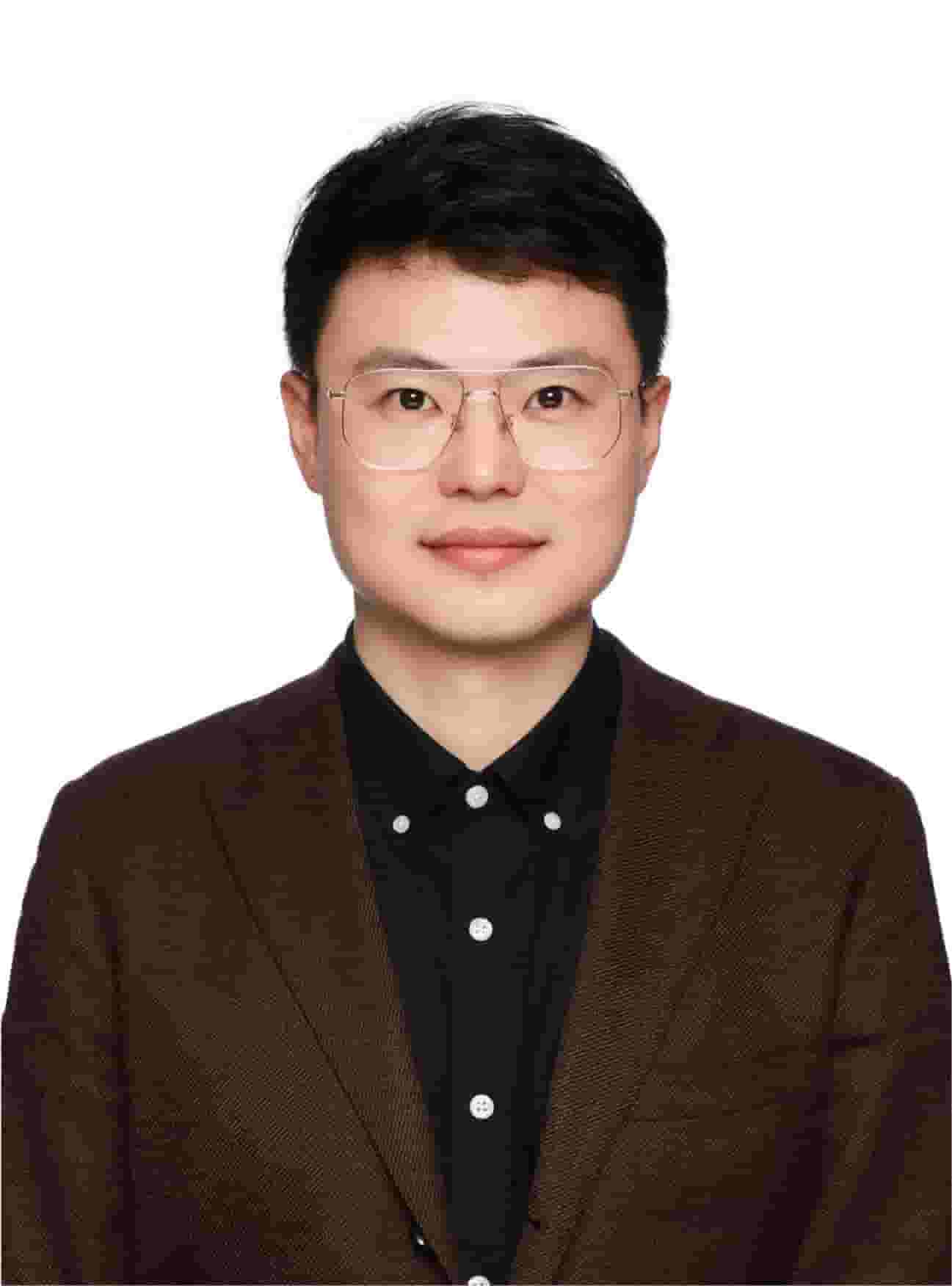}}]{Tianwei Zhang} is an Associate Research Scientist at Shenzhen Institute of Artificial Intelligence and Robotics for Society, The Chinese University of Hong Kong, Shenzhen. 
He received a doctoral degree from the Department of Mechano-Informatics, The University of Tokyo. He received a Master's degree in Electronic Science and Technology from Peking University in 2013.
His research interests are perceptive locomotion, humanoid robotics and dynamic SLAM. He has published more than 60 papers in these fields.
\end{IEEEbiography}%

\begin{IEEEbiography}[{\includegraphics[width=1in,
clip]{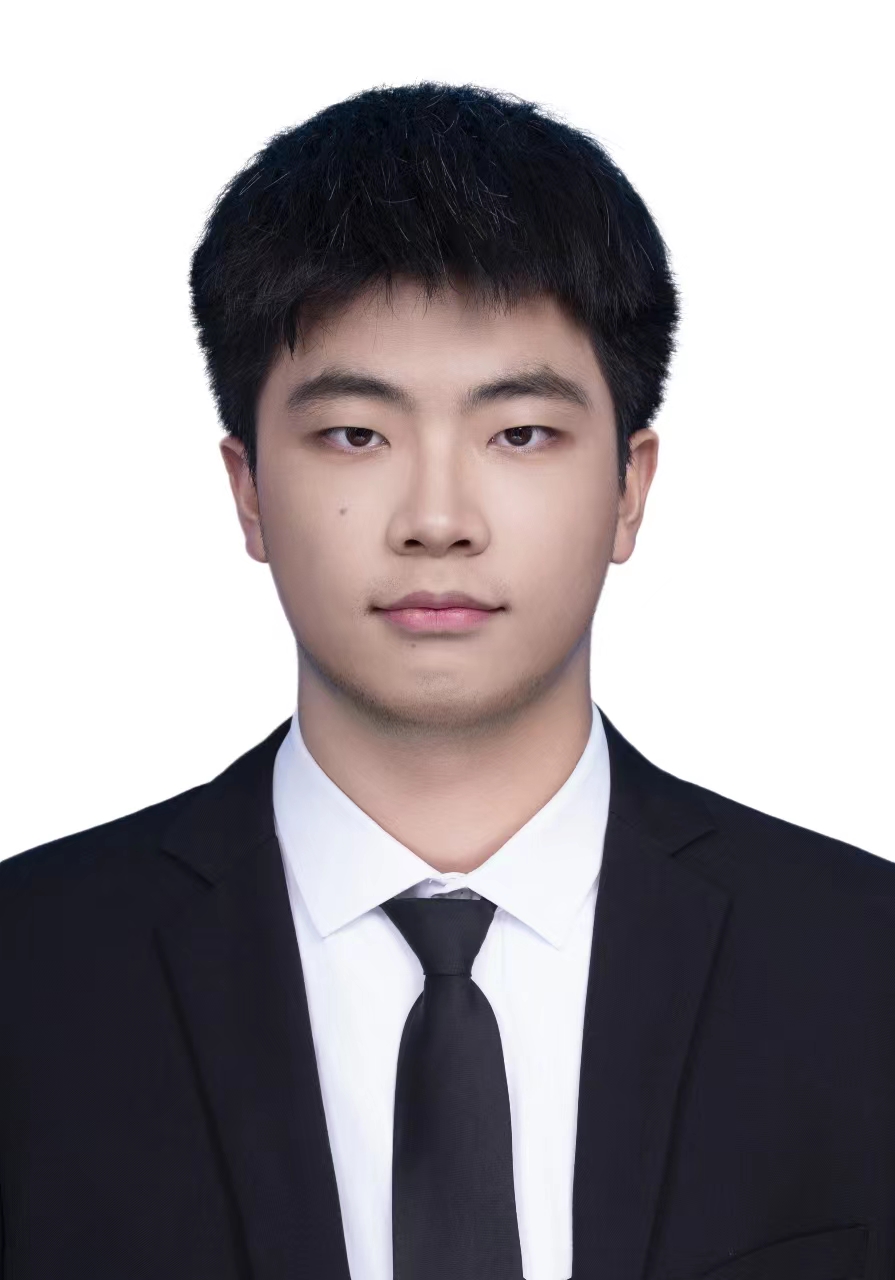}}]{Hanming Yan} received the B.E. degree in electronic information engineering from Nanjing Agricultural University, Nanjing, China. He is currently working toward the M.E. degree in instrumentation engineering with Shenzhen University, Shenzhen, China. His research interests include tactile sensors.
\end{IEEEbiography}%

\begin{IEEEbiography}[{\includegraphics[width=1in,
clip]{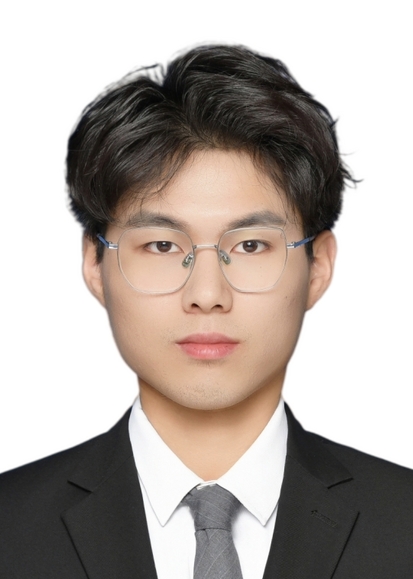}}]{Yang Yang} received the B.E. degree from Harbin Institute of Technology, Harbin, China, and the M.E. degree from Tsinghua Shenzhen International Graduate School, Shenzhen, China. His research interest is robotic manipulator control.
\end{IEEEbiography}%

\begin{IEEEbiography}[{\includegraphics[width=1in,
clip]{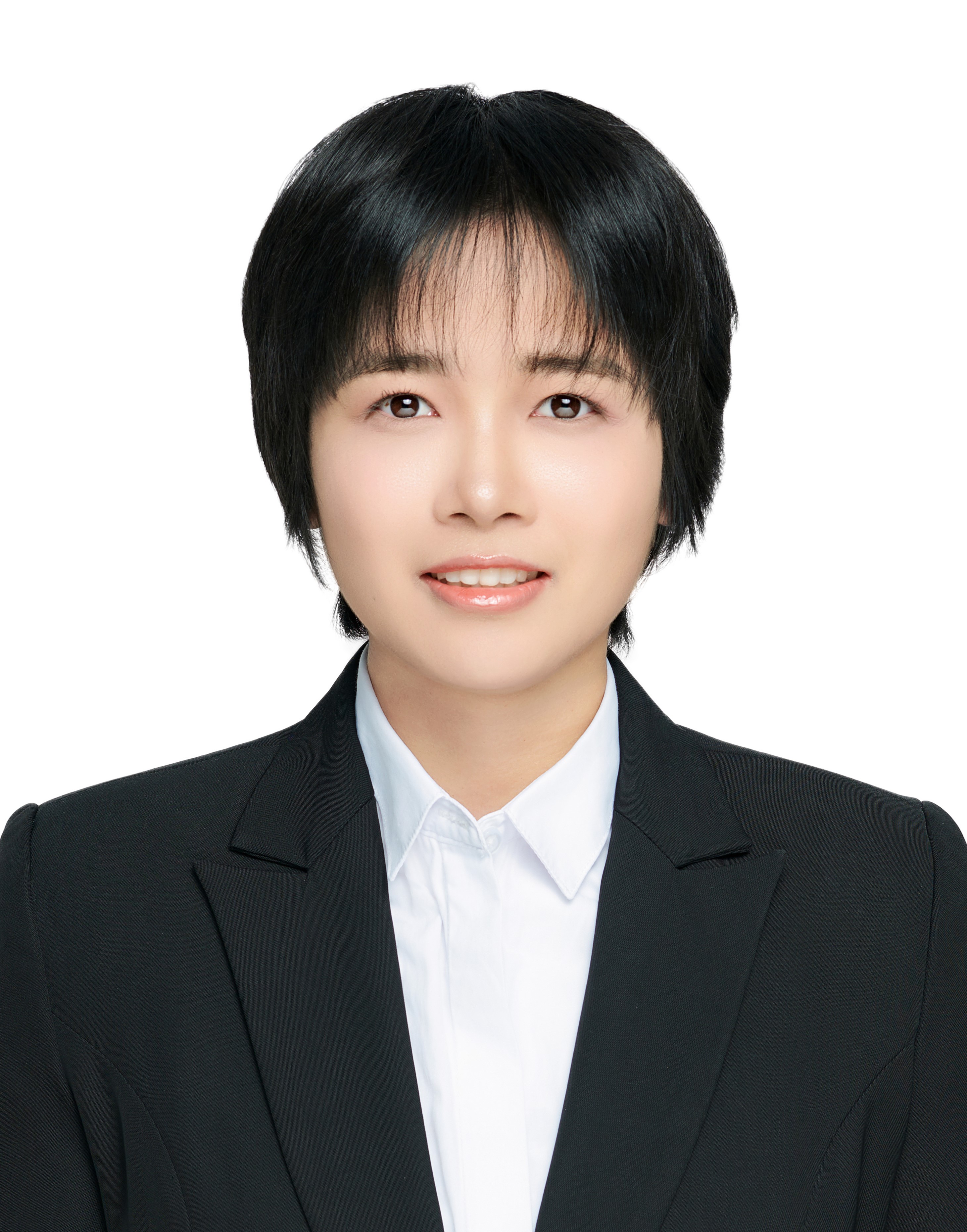}}]{Ziya Wang} is an Assistant Professor at Shenzhen University. She received both her PhD and bachelor’s degrees from the Department of Physics at the University of Science and Technology Beijing. Her research interests include tactile sensors, robotic tactile perception, and dexterous manipulation. She has published more than 50 papers in leading journals and international conferences in these fields.
\end{IEEEbiography}%
\end{CJK}
\end{document}